\documentclass{article}

\PassOptionsToPackage{numbers, compress}{natbib}
\usepackage[final]{neurips_2024}

\usepackage[utf8]{inputenc} % allow utf-8 input
\usepackage[T1]{fontenc}    % use 8-bit T1 fonts
\usepackage{hyperref}       % hyperlinks
\usepackage{url}            % simple URL typesetting
\usepackage{booktabs}       % professional-quality tables
\usepackage{amsfonts}       % blackboard math symbols
\usepackage{nicefrac}       % compact symbols for 1/2, etc.
\usepackage{microtype}      % microtypography
\usepackage[table]{xcolor}         % colors
\usepackage{amsmath}
\usepackage{amssymb}
\usepackage{array}
\usepackage{multirow}
\usepackage{caption}
\usepackage{subcaption}
\usepackage{graphicx}
\usepackage[ruled,vlined]{algorithm2e}
\usepackage{float}
\usepackage{wrapfig}
\usepackage{adjustbox}
\usepackage{enumitem}

\newcommand{\eqc}{\overset{c}{=}}
\newcommand{\method}{BMRS}
\newcommand{\methodn}{\method$_\mathcal{N}$}
\newcommand{\methodu}{\method$_\mathcal{U}$}

\usepackage[english]{babel}
\addto\captionsenglish{%
}
\addto\extrasenglish{%
}
\def \xbf{{\mathbf x}}
\def \wbf{{\mathbf w}}
\def \Wbf{{\mathbf W}}
\def \hbf{{\mathbf h}}

\title{BMRS: Bayesian Model Reduction for Structured Pruning}

\author{%
  Dustin Wright, Christian Igel, and Raghavendra Selvan \\
  Department of Computer Science, University of Copenhagen\\
  \texttt{\{dw,igel,raghav\}@di.ku.dk} \\
}

\begin{document}

\maketitle

\begin{abstract}

Modern neural networks are often massively overparameterized leading to high compute costs during training and at inference. One effective method to improve both the compute and energy efficiency of neural networks while maintaining good performance is structured pruning, where full network structures (e.g.~neurons or convolutional filters) that have limited impact on the model output are removed.
%One popular way to reduce the size of neural networks while maintaining good performance is to prune the weights which have limited impact on the predictive capabilities of the underlying models. Structured pruning does this at the level of full network structures (e.g. neurons or convolutional filters). This has benefits for efficiency over unstructured pruning which removes individual or subset of weights within network structures. 
%One line of work, which has its roots in variational dropout, tackles structured pruning from a Bayesian perspective by introducing random multiplicative noise on neural network structures with a sparsity inducing prior. 
%However, existing pruning methods which employ the Bayesian setup depend on selecting pruning thresholds which can achieve a balance of sparsity and performance. 
In this work, we propose Bayesian Model Reduction for Structured pruning (\method{}), a fully end-to-end Bayesian method of structured pruning. \method{} is based on two recent methods: Bayesian structured pruning with multiplicative noise, and Bayesian model reduction (BMR), a method which allows efficient comparison of Bayesian models under a change in prior. 
%We present \method{} as a framework for deriving different classes of structured pruning algorithms based on the choice of priors on the multiplicative noise operating on the weights. 
We present two realizations of \method{} derived from different priors which yield different structured pruning characteristics:  1) \methodn{} with the truncated log-normal prior, which offers reliable compression rates and accuracy without the need for tuning any thresholds and 2) \methodu{} with the truncated log-uniform prior that can achieve more aggressive compression based on the boundaries of truncation. %Both of the methods performed well  across datasets and models without requiring experiment-specific tuning of thresholds. 
Overall, we find that \method{} offers a theoretically grounded approach to structured pruning of neural networks yielding both high compression rates and accuracy. Experiments on multiple datasets and neural networks of varying complexity showed that the two \method{} methods offer a competitive performance-efficiency trade-off compared to other pruning methods.\footnote{Source code: \url{https://github.com/saintslab/bmrs-structured-pruning/}}%
% without the need for tuning experiment-specific pruning thresholds.
% 

%We show that the BMRS allows for high sparsification and accuracy without the need for tuning any thresholds for pruning. Additionally, we demonstrate that more extreme compression is achievable with We experimentally validate the pruning capabilities of these algorithms, showing that sparsity is induced in this framework by gradually increasing the precision of noise variables, and that pruning can be performed when the likelihood of a given noise variable to produce a floating point number at a desired level of precision is greater than random. Experiments show...
% \footnote{Source code: \url{https://anonymous.4open.science/r/bmrs-structured-pruning-8973/}}
\end{abstract}

\section{Introduction}
\label{sec:introduction}

Modern neural networks come with an increasing computational burden, as scale is often seen to be associated with performance~\cite{thompson2020computational}. The response to this has been a focus on research around the topic of neural network efficiency~\cite{bartoldson2023compute}, where the goal is to reduce the computational cost of a system while maintaining other desirable metrics. As such, selecting a method to improve efficiency comes with many tradeoffs, including how to balance compute and energy consumption with accuracy~\cite{DBLP:journals/corr/abs-2309-02065}.

Neural network pruning seeks to do this by removing parts of a network which have limited impact on its output. This comes in two primary forms: unstructured pruning, where individual weights are removed, and structured pruning, where entire neural network structures such as neurons and convolutional filters are removed~\cite{DBLP:journals/ijon/LiangGWSZ21}. Structured pruning is often desirable as unstructured pruning can result in sparse computations which are energy intensive on current hardware, while structured pruning can maintain more energy efficient dense operations~\cite{DBLP:journals/corr/abs-2002-05651,DBLP:conf/nips/JeonK18,peng2023efficiency}. Many ways to perform structured pruning have been proposed, but the challenge of how to appropriately balance accuracy and complexity in a principled manner has remained. 

In this work, we address this challenge by proposing \method{}: \textbf{B}ayesian \textbf{M}odel \textbf{R}eduction for \textbf{S}tructured pruning. \method{} is a principled method based on combining two complementary lines of work: Bayesian structured pruning with multiplicative noise~\cite{DBLP:conf/nips/NeklyudovMAV17} and Bayesian model reduction (BMR)~\cite{friston2018bayesian}, a method of efficient Bayesian model comparison under a change in prior. Multiplicative noise allows one to flexibly induce sparsity at any structural level without the need to use computationally complex spike-and-slab priors~\cite{DBLP:journals/nn/JantreBM23, DBLP:journals/corr/Markovic}, while BMR enables principled pruning rules without the need for task-specific threshold tuning. Starting with the approach from \cite{DBLP:conf/nips/NeklyudovMAV17}, we derive two versions of \method{} using different priors which offer their own benefits. \methodn{} is based on the truncated log-normal prior and has the benefit of achieving a high compression rate without needing to tune a threshold for compression, while \methodu{} offers tunable compression by controlling the allowable precision of noise variables in the network. In sum, our contributions are:
\begin{itemize}[leftmargin=*]
\setlength{\itemsep}{0pt}
    \item \method{}: a method for Bayesian structured pruning based on multiplicative noise and Bayesian model reduction;
    \item Derivations of pruning algorithms for two priors with theoretical motivation;
    \item Empirical results on a range of neural networks and datasets demonstrating high compression rates without any threshold tuning, with more extreme compression achievable via a parameter controlling allowable precision.
\end{itemize}

\begin{figure}[t]
        \centering
        \includegraphics[width=0.99\textwidth]{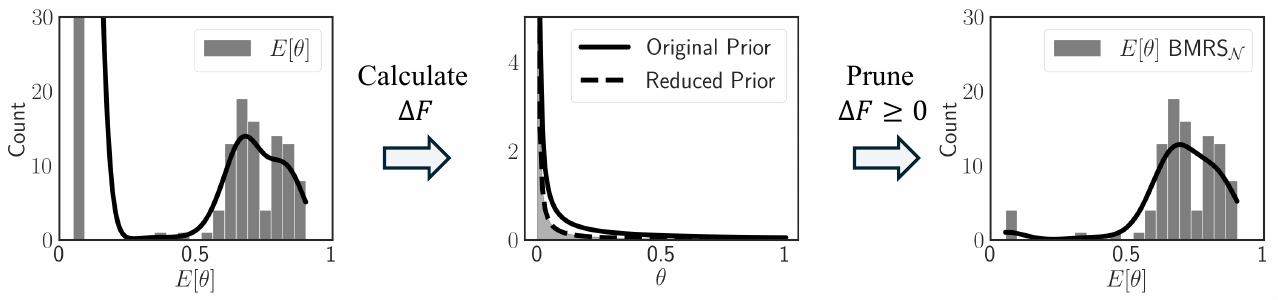}
        \caption{\method{} uses BMR to perform structured pruning under multiplicative noise by calculating the change in log-evidence of noise variables $\theta$ under a prior which would shrink them to 0.}%Plots are generated using a 5-layer MLP trained on CIFAR10.}
        %\caption{The expected values of learned noise variables $\theta$ follow an approximate spike-and-slab distribution after training. \method{} shrinks the original prior to a small region around 0, and determines the change in log-evidence under this new prior. Network weights which have greater log-evidence under the new prior are then pruned. Plots are generated using a 5-layer MLP trained on CIFAR10.}
        \label{fig:overview}
\end{figure}%

\section{Related work}

%{\bf Pruning}
The primary goal of neural network pruning is to determine the elements of a network which can be removed with minimal impact on the output. Ideally, a pruning method ranks all elements in the order in which they can be removed and provides a criterion for truncating the resulting ordered list.
%Pruning has its roots in the works of optimal brain damage~\cite{DBLP:conf/nips/CunDS89} and optimal brain surgeon~\cite{DBLP:conf/nips/HassibiS92}. These works showed that one can use first (gradient) and/or second-order (Hessian) information to determine the sensitivity of the output of a neural network with respect to its weights, and remove those weights which have lower sensitivity without significant reductions in accuracy. 
Since the early works on gradient based methods for pruning~\cite{DBLP:conf/nips/CunDS89,DBLP:conf/nips/HassibiS92}, the literature around neural network pruning has expanded rapidly, with the two main lines of work exploring pruning individual weights (unstructured pruning) and pruning full network structures (structured pruning). For a recent survey, see \cite{DBLP:journals/ijon/LiangGWSZ21}. The closest related works to ours are those pruning methods which perform Bayesian pruning~\cite{DBLP:conf/nips/NeklyudovMAV17,DBLP:conf/nips/LouizosUW17,DBLP:journals/jmlr/GhoshYD19,DBLP:journals/nn/JantreBM23,DBLP:journals/corr/abs-2402-11025}, and those which use Bayesian model reduction to determine what elements to remove from a neural network~\cite{beckers2024principled,DBLP:journals/corr/Markovic}. 

\paragraph{Bayesian pruning.}
Bayesian structured pruning was first explored in \citet{DBLP:conf/nips/BlumHP15}, where the authors demonstrate that dropout has a Bayesian interpretation as multiplicative noise with a sparsity inducing prior. The studies of \cite{DBLP:conf/nips/NeklyudovMAV17,DBLP:conf/nips/LouizosUW17,DBLP:journals/jmlr/GhoshYD19} follow this work by explicitly modeling the random noise in dropout with different priors,  \cite{DBLP:conf/nips/NeklyudovMAV17} using a truncated log-uniform prior and \cite{DBLP:conf/nips/LouizosUW17,DBLP:journals/jmlr/GhoshYD19} using horseshoe priors. Following this, the works of \cite{DBLP:conf/nips/BaiSC20,DBLP:journals/corr/abs-2305-00934,DBLP:journals/nn/JantreBM23,sun2022learning,DBLP:journals/corr/Markovic,DBLP:journals/corr/abs-2402-11025} have explored pruning of Bayesian neural networks (BNNs) with spike-and-slab priors to induce both weight sparsity and group sparsity with flat and hierarchical priors, respectively. \cite{DBLP:journals/nn/JantreBM23} demonstrate that thresholdless pruning is achievable by placing an explicit spike-and-slab prior on the nodes of a BNN to induce group sparsity. However, this setup requires complex and carefully constructed posteriors due to the discrete nature of spike-and-slab distributions and is thus computationally inefficient~\cite{DBLP:journals/corr/Markovic, DBLP:journals/nn/JantreBM23}. 
% Here, we show that thresholdless structured pruning is achievable efficiently through BMR with the multiplicative noise approach used in \cite{DBLP:conf/nips/NeklyudovMAV17} without the need to explicitly model spike-and-slab priors.

\paragraph{Bayesian model reduction.}
Bayesian model reduction, discussed in detail in \S\ref{sec:bmr}, is an efficient method of Bayesian model comparison which allows for analytic solutions for the model evidence under a change in priors. BMR has found application across multiple scientific disciplines~\cite{DBLP:journals/neco/FristonLFPHO17,DBLP:journals/neuroimage/FristonLORSWZZ16,kiebel2008dynamic}, and has recently been used as a method for neural network pruning~\cite{DBLP:journals/corr/Markovic,beckers2024principled}. More specifically, \cite{beckers2024principled} demonstrate the benefits of BMR-based pruning for the case of a BNN with a Gaussian prior 
%($p(w) = \mathcal{N}(w|0,1)$) and setting $\tilde{p}(w) = \mathcal{N}(w| 0,\epsilon)$ with sufficiently small $\epsilon$
on the weights, and \cite{DBLP:journals/corr/Markovic} demonstrate the utility of BMR for unstructured pruning of BNNs with priors inducing both weight and group sparsity. 

\section{Problem formulation}
\label{sec:background}

\subsection{Structured pruning with multiplicative noise and variational inference}
\label{sec:mult_noise}
We approach the problem of structured pruning using sparsity inducing multiplicative noise as 
described in \cite{DBLP:conf/nips/NeklyudovMAV17}. In this setting, we have a dataset consisting of $N$ i.i.d.{} input-output pairs, $\mathcal{D} = \{(\xbf_{j}, y_{j})~ \forall j=1,\dots,N\}$.
We consider a parametric model, here a deep neural network, that maps the input data $\xbf_j$ to their output $y_j$ using the trainable parameters $\Wbf$ giving rise to the likelihood function $p(\mathcal{D}|\Theta, \Wbf) = \prod_{j=1}^N p(y_j|\xbf_j,\Theta,\Wbf)$. In addition to the trainable weights, $\Wbf$, the model consists of the sparsity inducing multiplicative noise given by the random variable, $\Theta$, with prior $p(\Theta)$. This is in contrast to BNNs where the weights are random variables but aligns with the setting when using multiplicative noise for Bayesian pruning~\cite{DBLP:conf/nips/NeklyudovMAV17}.

The effect of the multiplicative noise $\theta_i \in \Theta$ for a structural element in a neural network with index $i$, parameters $\wbf_i$, and input $\hbf_{i-1}$  is given as
\begin{equation}
    \hbf_{i} = \theta_{i} \cdot ({\wbf}_i \mathbf{h}_{i-1}) \quad,\quad \theta_i \sim p(\theta_i).
\end{equation}

We note that $\wbf_{i}$ could be the parameters of any structural element in the network, for example, a single neuron or an entire convolutional filter. Given this, we would like to learn  the maximum likelihood estimate (MLE) $\hat{\wbf_i}$ of the weights
%in a supervised setting using $\mathcal{D}$, 
as well as the posterior distribution over the multiplicative noise, $p(\theta_{i} | \mathcal{D},\hat{\wbf_i})$, when using a sparsity inducing prior $p(\theta_{i})$ such that $\theta_{i}$ favors values closer to {0}. 

Following~\cite{DBLP:conf/nips/NeklyudovMAV17}, the neural network weights are learned via gradient descent as in standard deep learning model optimization.
%, $\hat\wbf_i = \arg\max_{\wbf_i} p(\mathcal{D}|\theta_i, \wbf_i)$. 
%In case of post-training pruning, $\hat{\Wbf}$ corresponds to a converged model, whereas in continuous pruning, $\hat{\Wbf}$ denotes the updated model weights after each gradient update within a training epoch.
The posterior distribution, $p(\theta|\mathcal{D},\hat{\wbf_i}) = p(\mathcal{D}|\theta_i,\hat{\wbf}_i) p(\theta_i)/p(\mathcal{D})$, however, is intractable. We resort to a variational approximation from a tractable family of approximating distributions, $q_{\mathbf{\phi}}(\theta)$, parameterized by $\mathbf{\phi}$ (for the sake of brevity we do not indicate the dependence on $\mathbf{w}_i$ and omit the subscript $i$). The parameters $\phi$ are obtained by optimizing the following objective w.r.t.{} $\theta$:
\begin{equation}
\label{eq:elbo}
    \text{F}[p,q] = 
    D_{\text{KL}}[q_{\mathbf{\phi}}(\theta) || p(\theta | \mathcal{D})] \eqc D_{\text{KL}}[q_{\mathbf{\phi}}(\theta) || p(\theta)] - \mathbb{E}_{q_{\mathbf{\phi}}}[\log p(y_{j}|\xbf_{j}, \theta, \hat{\wbf})]
\end{equation}
This is the commonly used variational free energy (VFE) or negative evidence lower bound (ELBO)~\cite{beal2003variational}. Here $\eqc$ denotes equality up to a positive constant.

The expectation $\mathbb{E}_{q_{\mathbf{\phi}}}[\cdot]$ is approximated by a Monte Carlo estimator acting on minibatch samples from $\mathcal{D}$, and reparameterization allows to backpropagate gradients through stochastic variables~\cite{DBLP:journals/corr/KingmaW13,DBLP:conf/nips/LouizosUW17}. Under this reparameterization, the variational distribution, $q_{\mathbf{\phi}}(\theta)$, becomes a deterministic function of the non-parametric noise $\epsilon \sim p(\epsilon)$ and the VFE is calculated as
\begin{equation}
    \text{F}[p,q] \eqc D_{\text{KL}}[q_{\mathbf{\phi}}(\theta) || p(\theta)] - \sum_{(x_{j}, y_{j}) \in \mathcal{D}}\log p(y_{j}|x_{j}, \theta = f(\mathbf{\phi}, \epsilon) ; \hat{\wbf}),
\label{eq:svi_nn}
\end{equation}
where $f$ is a function that allows us to sample $\theta$ via deterministic parameters $\mathbf{\phi}$ and the non-parametric stochastic variable, $\epsilon$. Optimization of \autoref{eq:svi_nn} allows us to jointly learn $\hat{\wbf}$ and $\theta$. The particular choice of priors and the approximating distributions to induce sparsity are discussed in \S\ref{sec:noise_layer}.

\subsection{Bayesian model reduction}\label{sec:bmr}

Bayesian model reduction (BMR) allows one to \textit{efficiently} compute the change in VFE (\autoref{eq:elbo}) under a change in prior without the need to re-estimate model parameters. To perform pruning, one can start out by selecting a broad prior for the original model estimation and then pick a narrower prior (i.e. reduced prior) with the density concentrated around 0.
%for the new prior, and use 
Then, BMR can be used to determine if the VFE is greater under the reduced model, and prune those parameters for which this condition holds. We briefly describe how this is achieved in the general case, followed by the specific realization for \method{} in \S\ref{sec:bmrs}; for further details see \cite{friston2018bayesian}.

Consider the likelihood function $p(\mathcal{D}|\theta)$ and a prior $p(\theta)$ on the variable $\theta$. We can introduce a new prior $\tilde{p}(\theta)$ which shares the same likelihood as the original model (i.e. $p(\mathcal{D}|\theta) = \tilde{p}(\mathcal{D}|\theta)$) and get: 
%With this, we can use Bayes rule to see the following equivalence:
\begin{equation}
\begin{split}
    p(\mathcal{D}|\theta) = \frac{p(\theta|\mathcal{D})p(\mathcal{D})}{p(\theta)} = \frac{\tilde{p}(\theta|\mathcal{D})\tilde{p}(\mathcal{D})}{\tilde{p}(\theta)} \Rightarrow
    \tilde{p}(\theta|\mathcal{D}) = \frac{p(\mathcal{D})}{\tilde{p}(\mathcal{D})}p(\theta|\mathcal{D})\frac{\tilde{p}(\theta)}{p(\theta)}
\end{split}
\end{equation}
By marginalizing over $\theta$ and taking the log, we  obtain the difference in log evidence as:
\begin{equation}
\begin{split}
    \log \tilde{p}(\mathcal{D}) - \log p(\mathcal{D}) = \log \int p(\theta|\mathcal{D})\frac{\tilde{p}(\theta)}{p(\theta)}d\theta 
    \approx \log \int q_{\mathbf{\phi}}(\theta)\frac{\tilde{p}(\theta)}{p(\theta)}d\theta = \log \mathbb{E}_{\tilde{p}}\left[\frac{q_{\mathbf{\phi}}(\theta)}{p(\theta)}\right]
\end{split}
\label{eq:dF_derivation}
\end{equation}
More concisely, we  call the change in log evidence $\Delta F$ and thus have:
\begin{equation}
    \Delta F \triangleq \log \tilde{p}(\mathcal{D}) - \log p(\mathcal{D}) \approx  \log \mathbb{E}_{\tilde{p}}\left[\frac{q_{\mathbf{\phi}}(\theta)}{p(\theta)}\right]
\label{eq:dF}
\end{equation}
If the new prior, $\tilde{p}(\theta)$, is selected so that $\theta$ would be removed, pruning can be performed when $\Delta F \ge 0$. Additionally, when the type of distributions between $p(\theta)$, $\tilde{p}(\theta)$, and $q_{\mathbf{\phi}}(\theta)$ are the same or similar (e.g. Gaussian), $\Delta F$ can be calculated efficiently in closed form (see \cite{DBLP:journals/neuroimage/FristonP11}). 

We presented BMR  for a general likelihood function $p(\mathcal{D}|\theta)$; it holds analogously for the likelihood function $p(\mathcal{D}|\theta,\Wbf)$ introduced with the multiplicative noise described in \S\ref{sec:mult_noise}.

% 

% \input{problem_formulation_DW}

%In order to use BMR for pruning, one can first train a model with a very broad or uninformative prior and then find $\Delta F$ when using a prior which would shrink the parameter $\theta$ to 0. Then, if the model under the new prior provides a better explanation for the data (i.e. $\Delta F > 0$), one can prune that parameter.

\section{Bayesian model reduction for structured pruning (BMRS)}
\label{sec:full_method}
% \begin{figure}[t]
%         \centering
%         \includegraphics[width=0.98\textwidth]{figures/figure1_annotated.pdf}
%         \caption{The expected values of learned noise variables $\theta$ follow a spike-and-slab distribution after training. \method{} shrinks the original prior to a small region around 0, and determines the change in log-evidence under this new prior. Neurons which have greater log-evidence under the new prior are then pruned. Plots are generated using a 5-layer MLP trained on CIFAR10. }
% \end{figure}%

% Inspired by this, we show how BMR can be used to derive multiple principled pruning algorithms and enable high sparsification for structured pruning without the need to tune pruning thresholds.

Our goal is to derive a principled structured pruning algorithm starting from the general formulation in \autoref{sec:background} which can automatically determine which structures to prune. BMRS accomplishes this by following the multiplicative noise setup in~\cite{DBLP:conf/nips/NeklyudovMAV17} with BMR used on the noise terms. Figure~\ref{fig:overview} illustrates the general approach, where $\Delta F$ is calculated for a model trained with multiplicative noise under a reduced prior, and elements of the model are removed if the new VFE is greater.  We next describe the multiplicative noise layer trained using \autoref{eq:elbo}, and then derive two variants of \method{} from \autoref{eq:dF} using  different reduced priors.

\subsection{Multiplicative noise layer}
\label{sec:noise_layer}
%A popular method used to regularize a neural network is dropout~\cite{}, where neurons are randomly perturbed or dropped completely during training. Dropout can be viewed as applying multiplicative noise to a given neuron, i.e. $\hat{h}_{i} = \theta * h_{i}$ where $h_{i}$ is the value of a particular neuron and $\theta$ is a random variable drawn from some distribution $p(\theta)$. \cite{DBLP:conf/nips/BlumHP15} show that when viewed in this way, training with dropout induces an implicit Bayesian neural network learned with variational inference. In the case wherweightse $\theta$ is Gaussian with $p(\theta) = \mathcal{N}(1, \alpha)$, a model with weights $\mathbf{W}$ learns a posterior distribution over $w_{i,j} \in \mathbf{W}$ as $q_{\mathbf{\phi}}(w_{i,j}) = \mathcal{N}(\omega_{i,j}, \alpha \omega_{i,j}^{2})$, where $\omega_{i,j}$ denotes the learnable parameter of $w_{i,j}$. Training the network involves maximizing the expected log-likelihood $\mathcal{L}_{D}$ on some dataset $D$, equivalent to maximizing $\mathbb{E}_{q}[\log p(D|\theta)]$ with an implicit regularizer $\text{KL}[q_{\mathbf{\phi}}(\theta) || p(\theta)]$ and thus satisfying \autoref{eq:elbo}. The induced KL divergence term requires the existence of an appropriate prior, which  \cite{DBLP:conf/nips/BlumHP15} find to be the scale invariant log-uniform prior i.e. $p(\log |w_{i,j}|) \propto c$.

The concept of multiplicative noise inducing sparsity in neural networks was first introduced with variational dropout, where \cite{DBLP:conf/nips/BlumHP15} show that dropout has a Bayesian interpretation as multiplicative noise with a log-uniform prior. One can use this interpretation of dropout in order to explicitly learn dropout parameters, $\theta_{i}$, as in \S\ref{sec:background}, by selecting appropriate prior and variational distributions and optimizing \autoref{eq:elbo} directly. \cite{DBLP:conf/nips/NeklyudovMAV17} propose to do so by using the truncated log-uniform distribution as a prior and the truncated log-normal distribution as the variational distribution. 
%However, the log-uniform distribution is an improper prior (i.e. unnormalized with infinite support), so the KL divergence is intractable and must be approximated due to the prior gap. To address this, they propose to use truncated distributions, which makes the KL divergence calculable in closed form. 
As such, the variational approximation can be performed using 
\begin{equation}
    \hbf_{i} = \theta_{i} \cdot (\mathbf{w}_{i}\mathbf{h}_{i-1}); \quad q_{\mathbf{\phi}}(\theta_{i}) = \text{LogN}_{[a,b]}(\theta_{i}| \mu_{i}, \sigma^{2}_{i}); \quad p(\theta_{i}) = \text{LogU}_{[a,b]}(\theta_{i})
\end{equation}
with bounded support between $a$ and $b$ and $0 < a < b \le 1$. We refer to \cite{DBLP:conf/nips/NeklyudovMAV17} for details on how to learn $q_{\mathbf{\phi}}$, which is obtained by optimizing \autoref{eq:svi_nn}.
%\autoref{sec:lmn-dropout-details}.  

The log-uniform distribution serves as a sparsity inducing prior as most of its density is concentrated around 0 (see panel 2 in \autoref{fig:overview}). Additionally, it acts as a regularizer on the floating point precision of the multiplicative noise terms~\cite{DBLP:conf/nips/BlumHP15}. In \cite{DBLP:conf/nips/NeklyudovMAV17} this is used to perform structured pruning by removing all structures $\hbf_{i}$ where the signal-to-noise ratio of the noise term $\theta_{i}$ falls below a pre-defined threshold. We next show how to derive principled pruning algorithms based on BMR which induce sparsity while maintaining accuracy without the need for tuning pruning thresholds.

\subsection{Deriving \method{}}
\label{sec:bmrs}

 Our goal is to use BMR in order to perform structured pruning of models trained with multiplicative noise. To do so, we must select a new prior $\tilde{p}(\theta)$ from which we can: 1) induce sparsity; 2) efficiently calculate $\Delta F$ and; 3) prune the network while maintaining good performance.
 
Selecting the reduced prior, $\tilde{p}(\theta)$, is straightforward when the prior and approximate posterior are the same type of distributions. For example, in a fully BNN where one assumes a prior distribution of $\mathcal{N}(\Theta|{\bf 0}, \mathbf{I})$ over all the model weights with a mean-field variational approximation resulting in a factorisation over individual weights, $\mathcal{N}(\theta| \mu, \sigma^{2})$, the three criteria above can be met when one selects a Gaussian reduced prior with slight variance around 0 i.e. $\mathcal{N}(\theta| 0, \epsilon), \epsilon \approx 0$.\footnote{It can be shown that $\Delta F$ is calculable efficiently in closed form for this setup; see e.g. \cite{DBLP:journals/neuroimage/FristonP11}} However, in the case of multiplicative noise, our prior and variational distributions are of different types and thus the selection of the reduced prior is not immediately obvious. Here, we derive and compare the characteristics of two different reduced priors: one based on a truncated log-normal distribution, which we can use to approximate a Dirac delta at 0 (\methodn{}), and one based on a truncated log-uniform distribution with reduced support (\methodu{}).

 \subsubsection{BMRS with log-normal reduced prior (\methodn{})}

First, we derive $\Delta F$ when using the log-uniform distribution as the original prior, $p(\theta)=\text{LogU}_{[a,b]}(\theta)$, and the truncated log-normal distribution as the reduced prior, $\tilde{p}(\theta) = \text{LogN}_{[a,b]}(\theta|\tilde{\mu}_{p}, \tilde{\sigma}_{p}^{2})$. We select a truncated log-normal distribution, as it matches the variational distribution $q_{\mathbf{\phi}}(\theta)$, and the log-uniform prior, because it is a special case of the log-normal distribution when the variance goes to infinity. Because of this, we expect that $\Delta F$ will have a closed form solution, and that the computation will be efficient. We briefly present the results of the derivation here; for the full derivation see \autoref{sec:BMR_deriv}.

We can use the specific forms of $p(\theta)$ and $\tilde{p}(\theta)$ for the truncated log-uniform and truncated log-normal distributions, respectively, in \autoref{eq:dF} to determine $\Delta F$:
\begin{equation}
    \Delta F \approx  \log \mathbb{E}_{\tilde{p}}\left[\frac{q_{\mathbf{\phi}}(\theta)}{p(\theta)}\right] = 
    \log\frac{Z_{\tilde{q}}(\log b - \log a)}{Z_{\tilde{p}}Z_{q}} + \frac{1}{2}\log\frac{\tilde{\sigma}_{q}^{2}}{2\pi\tilde{\sigma}_{p}^{2}\sigma_{q}^{2}} - \frac{1}{2}\left(\frac{\mu_{q}^{2}}{\sigma_{q}^{2}} + \frac{\tilde{\mu}_{p}^{2}}{\tilde{\sigma}_{p}^{2}} - \frac{\tilde{\mu}_{q}^{2}}{\tilde{\sigma}_{q}^{2}}\right)
\label{eq:dF_logn}
\end{equation}
with $\displaystyle \tilde{\sigma}_{q}^{2} = \left(\frac{1}{\sigma_{q}^{2}} + \frac{1}{\tilde{\sigma}_{p}^{2}}\right)^{-1} \text{ and }
\tilde{\mu}_{q} = \tilde{\sigma}_{q}^{2}\left(\frac{\mu_{q}}{\sigma_{q}^{2}} + \frac{\tilde{\mu}_{p}}{\tilde{\sigma}_{p}^{2}}\right)$.  Here, $Z_{p} = \Phi(\beta_{p}) - \Phi(\alpha_{p})$; $\Phi(t) = \frac{1}{2}[1 + \text{erf}(\frac{t}{\sqrt{2}})]$ is the CDF of the standard Normal distribution, $t\sim \mathcal{N}(0,1)$; $\alpha_{p} = ({a - \mu_{p}})/{\sigma_{p}}$; and $\beta_{p} = ({b - \mu_{p})/}{\sigma_{p}}$.

As can be seen from \autoref{eq:dF_logn}, the calculation for $\Delta F$ can be performed directly using only the statistics of the priors and variational distribution. In order for \autoref{eq:dF_logn} to induce sparsity, we must select a $\tilde{\mu}_{p}$ and $\tilde{\sigma}_{p}^{2}$ that effectively collapse $\theta$ to 0. To achieve this, we can approximate a Dirac delta at 0 by selecting $\tilde{\mu}_{p}$ to be close to 0 (e.g., the lower bound of truncation $a$), and $\tilde{\sigma}_{p}^{2}$ to be sufficiently small. We will demonstrate in \S\ref{sec:experiments} that this reduced prior results in high sparsity while maintaining performance without any need for tuning pruning thresholds. %as we simply select the prior to approximate a Dirac delta at 0.

%In practice, the selection of $\tilde{\mu}_{p}$ allows us some degree of control over the compression rate, with higher $\tilde{\mu}_{p}$ leading to higher compression. We relate this to the connection between the log-uniform prior and floating point format for storing numbers \cite{DBLP:conf/nips/BlumHP15,hamming1970distribution}. From the perspective of the minimum description length (MDL) principle, the standard IEEE floating point format is close to optimal for encoding numbers when the decoder has a log-uniform prior with a fixed support. In other words, floating-point numbers drawn at random in floating point format closely follow a log-uniform distribution. Given this, the KL-divergence optimized in \autoref{eq:elbo} acts as a regularizer on the number of significant digits required by a given parameter (in practice driving $\mu_{q}$ down and $\sigma_{q}^{2}$ up). The selection of $\tilde{\mu}_{p}$ and subsequent calculation of $\Delta F$ then effectively allows us to determine if a given parameter is more likely to have been drawn at a fixed precision between $\log a$ and $\log b$.  This becomes clearer for the case of exact priors, which we discuss next. 

\subsubsection{BMRS with log-uniform  reduced prior (\methodu{})} 

Next, we derive the change in VFE, $\Delta F$, when using a truncated log-uniform distribution as the original prior, ${p}(\theta) = \text{LogU}_{[a, b]}(\theta)$, and a truncated log-uniform distribution with reduced support as the reduced prior, $\tilde{p}(\theta) = \text{LogU}_{[a', b']}(\theta)$. We select a reduced truncated log-uniform distribution for the same reasons as the truncated log-normal: we expect that $\Delta F$ will have an efficiently calculable closed form, given that the priors are of the same type and are a special case of the variational distribution. The PDF of the reduced truncated log-uniform distribution is given as follows:
\begin{equation}
    \tilde{p}(\theta) = \text{LogU}_{[a', b']}(\theta) = \begin{cases}
    \left({\theta\log \frac{b'}{a'}}\right)^{-1}, & a < a' \le \theta \le b' < b \\
    0, & \text{otherwise}
  \end{cases}
\end{equation}
Using this, we can directly solve the integral under the expectation given in \autoref{eq:dF} for $\Delta F$ (full details in \autoref{sec:bmrsu_derivation}):
\begin{equation}
    \exp \Delta F \approx \mathbb{E}_{\tilde{p}}\left[\frac{q_{\mathbf{\phi}}(\theta)}{p(\theta)}\right] = \int_{a}^{b}\text{LogU}_{[a',b']}(\theta)\frac{q_\mathbf{\phi}(\theta)}{\text{LogU}_{[a,b]}(\theta)} d\theta
%\end{equation*}
%\begin{equation*}
%    = \int_{a}^{a'}0d\theta + \int_{a'}^{b'}\frac{\log\frac{b}{a}}{\log \frac{b'}{a'}}q_\mathbf{\phi}(\theta)d\theta + \int_{b'}^{b}0d\theta 
= \frac{\log\frac{b}{a}}{\log \frac{b'}{a'}}q_{\mathbf{\phi}}(a' \le \theta_{i} \le b')
\end{equation}
% \begin{equation*}
%     \Phi_{q}(a', b') = \frac{\Phi(\frac{b' - \mu_{q}}{\sigma_{q}}) - \Phi(\frac{a' - \mu_{q}}{\sigma_{q}})}{Z_{q}}
% \end{equation*}
where $q_{\mathbf{\phi}}(a' \le \theta_{i} \le b')$ is the CDF of the variational distribution evaluated between $a'$ and $b'$. We know from \autoref{eq:dF_derivation} that the VFE under $\tilde{p}(\theta)$ is greater when $\exp \Delta F \ge 1$. Plugging this in:
\begin{equation}
    1 \le \frac{\log\frac{b}{a}}{\log \frac{b'}{a'}}q_{\mathbf{\phi}}(a' \le \theta_{i} \le b')
\Rightarrow
\label{eq:exact_df}
    \frac{\log\frac{b'}{a'}}{\log \frac{b}{a}} \le q_{\mathbf{\phi}}(a' \le \theta \le b')
\end{equation}
Here, the left hand side of the inequality is the CDF of the truncated log-uniform distribution between $a'$ and $b'$. In other words, when the new prior is a log-uniform distribution with reduced support, the BMR pruning criterion amounts to a comparison between the CDF of the original prior and the variational distribution along the interval $[a', b']$. Additionally, \autoref{eq:exact_df} shows that this is generalizable to any variational distribution with support broader than the reduced prior.
\begin{algorithm}[t]
\caption{Training and pruning with \method{}\label{alg:training}}
\SetKwInOut{Input}{input}
\SetArgSty{textnormal}
\DontPrintSemicolon
\Input{dataset $\mathcal{D}$; neural network with deterministic weights $\mathbf{W}$ and variational parameters $\mathbf{\phi}$;  original prior p($\Theta$); reduced prior $\tilde{p}(\Theta)$; number of training epochs $e_{T}$; number of fine-tuning epochs $e_{F}$; number of pruning epochs $P$}
$j \gets 0$\;
\While{$j < e_{T}$ and $\mathbf{W}, \mathbf{\phi}$ not converged}{
    Train $\mathbf{\phi}$ and $\mathbf{W}$ on $\mathcal{D}$ using \autoref{eq:svi_nn}\;

    \If{$j \bmod P = 0$}{
        \ForAll{$\theta_{i}$ in $\Theta$}{
        dF $\gets \Delta F(p(\theta_{i}), \tilde{p}(\theta_{i}), q_{\mathbf{\phi}}(\theta_{i}))$\;
            \If{dF $\ge 0$}{
                $\mathbf{\phi} \gets \mathbf{\phi} \setminus \mathbf{\phi}_{i}$\;
                $\mathbf{W} \gets \mathbf{W} \setminus \mathbf{w}_{i}$
            }
        }
    }
    $j \to j + 1$
}
Fine tune $\mathbf{W}$ and $\mathbf{\phi}$ on $\mathcal{D}$ for $e_{F}$ epochs
\end{algorithm}

{\bf \methodu{} pruning and connection to floating point precision.} To see how \methodu{} can be used for pruning, we first briefly summarize the relationship between the log-uniform distribution and floating point numbers. Floating point numbers are commonly encoded in binary as a combination of a sign bit $s$, a set of exponent bits $e$, and a set of mantissa bits $m$ denoting the fractional part of a real-number: $r = s\cdot(m / 2^{p-1})\cdot2^{e}$, where $p$ determines the precision of the encoding. As discussed in \cite{hamming1970distribution}, 
the mantissae of ``naturally observed'' floating point numbers (e.g., natural constants)
tend to follow a log-uniform distribution and 
repeated multiplications/divisions on a digital computer transform a broad class of distributions towards a log-uniform distribution 
\begin{equation}
    m \sim \left({m\log B}\right)^{-1}, \quad {1}/{B} \le m \le 1,
\end{equation}
where $B$ is the base of the number system. In the case where the mantissa uses $p$ bits, there are $2^p$ representable fractional numbers so $B = 2^{p}$. As such, $p$ determines the smallest fractional value which can be represented. We can use this to have the reduced prior cover a finite range of high precision values which are acceptable to prune. To accomplish this, we use a reduced log-uniform prior of the following form by selecting two integers $p_{1}, p_{2}$ where $0 \le p_{1} < p_{2}$:
\begin{equation}
    \tilde{p}(\theta) = \begin{cases}\left({\theta\log\frac{2^{p_{2}}}{2^{p_{1}}}}\right)^{-1}, & {1}/{2^{p_{2}}} \le \theta \le {1}/{2^{p_{1}}} \\
    0, & \text{otherwise}
    \end{cases}
\end{equation}
Thus we can reduce the prior to a range of precision between $p_{1}$ and $p_{2}$ by selecting $a' = {1}/{2^{p_{2}}}$ and $b' = {1}/{2^{p_{1}}}$. \autoref{eq:exact_df} then has a natural interpretation as comparing the probability of drawing a random mantissa from the interval $[{1}/{2^{p_{2}}}, {1}/{2^{p_{1}}}]$ to the probability of drawing a value within that interval from the variational distribution. If we select $p_{2}$ to be the limit of the precision of values in the number system used ($p_{2}=23$ for single-point precision), then we can interpret $p_{1}$ as determining the prunable range of precision of all variables $\Theta$. The accuracy-complexity tradeoff inherent in the selection of $\tilde{p}(\theta)$ is then controlled through $p_{1}$, the desired cutoff of the precision of the network.

\subsection{Training and pruning}

The details on how to train a network and use \method{} for pruning are given in Algorithm \ref{alg:training}. We train a model for a fixed number of epochs (or until convergence) and perform pruning every $P$ epochs. This lends itself to either post-training pruning, where the network is fully trained followed by pruning and fine-tuning, or continuous pruning, where pruning is performed during model training. In our experiments, we explore both of these setups and contrast \method{} with alternative pruning methods. 

%TODO: Another interpretation: The selection of $\text{LogU}_{[a',b']}(\theta)$ for $\tilde{p}(\theta)$ can maybe be seen as trying to communicate messages $m \sim  q_{\mathbf{\phi}}(\theta)$ over a noisy channel which limits their precision. If we have a greater than random chance of decoding those messages, then we can say the evidence under the new prior is greater than the evidence under the old prior (need to think about how to make this connection stronger).

\section{Experiments}
\label{sec:experiments}
We demonstrate the pruning behavior of \method{} through several experiments with neural networks of varying complexity measured as the number of trainable parameters. 
%The datasets which we use in our experiments are as follows:
%\paragraph{MNIST \cite{}} ...
%\paragraph{Fashion-MNIST \cite{}} ...
%\paragraph{CIFAR10 \cite{}} ...
%\paragraph{TinyImagenet \cite{}} ...
We use the following datasets (full details in \autoref{sec:dataset_details}): MNIST \cite{lecun1998mnist}, Fashion-MNIST \cite{DBLP:journals/corr/abs-1708-07747}, CIFAR10 \cite{krizhevsky2009learning}, and TinyImagenet.
% \footnote{\url{https://huggingface.co/datasets/zh-plus/tiny-imagenet}}\footnote{For dataset details, see \autoref{sec:dataset_details}} 
For MNIST, Fashion-MNIST, and CIFAR10 we experiment with both a multi-layer perceptron (MLP) and a small CNN (Lenet5 \cite{DBLP:journals/pieee/LeCunBBH98}). Pruning layers are applied after each fully connected layer for the MLP, and for each convolutional filter and fully connected layer for Lenet5. For CIFAR10 and TinyImagenet, we further experiment with a pretrained ResNet-50 \cite{DBLP:conf/cvpr/HeZRS16} and a pretrained vision transformer (ViT) \cite{wu2020visual}. For ResNet-50, we apply pruning layers after each layer of batch normalization, and for ViT we apply pruning layers to the output of each transformer block. For the multiplicative noise layers, we set the left bound of truncation to be $\log a = -20$ and the right bound of truncation to be $\log b = 0$. Hyperparameters for the MLPs and Lenet5 are tuned on a model with no pruning performed and kept the same for each variant (see \autoref{sec:model_details}). 
%We perform experiments with \methodn{} and \methodu{} and compare to two baselines using the same multiplicative noise layers:
We perform experiments using the following model variants and baselines which cover both Bayesian pruning criteria (e.g. ours, SNR, and ${\mathbb{E}_{q_{\mathbf{\phi}}}[\theta]}$) as well as magnitude pruning (L2):
\begin{itemize}[leftmargin=*]
\setlength{\itemsep}{0pt}
    \item {\bf None:} A baseline with no compression and no multiplicative noise.
    \item {\bf L2:} A magnitude pruning baseline based on the L2 Norm of weight vectors (matrices in the case of convolutional filters) at the input of each neuron to be pruned~\cite{DBLP:conf/iclr/0022KDSG17}. We set the pruning threshold to the compression rate achieved by \methodn{} using the same settings of a given experiment.
    \item {\bf ${\mathbb{E}_{q_{\mathbf{\phi}}}[\theta]}$}: The expected value of noise variables $\theta$. For continuous pruning, we use a set threshold of 0.1.
    \item {\bf SNR:} The signal-to-noise ratio ${\mathbb{E}_{q_{\mathbf{\phi}}}[\theta]}\big/{\sqrt{\text{Var}[\theta]}}$ as used in \cite{DBLP:conf/nips/NeklyudovMAV17}. For continuous pruning, we use a set threshold of 1 as recommended in \cite{DBLP:conf/nips/NeklyudovMAV17}.
    \item {\bf \methodn{}:} \method{} using the log-normal prior from \autoref{eq:dF}. In order to reduce the prior to 0, we set $\tilde{\mu}_{p}$ to the left bound of truncation ($a$), and $\tilde{\sigma}^{2}_{p}$ to $10^{-12}$.
    \item {\bf \methodu{}-$p_{1}$:} \method{} using the log-uniform prior from \autoref{eq:exact_df}. In our experiments, we set $a'$ to be the limit of the precision of single-point floats ($p_{2}$ = 23 so $a' = {1}/{2^{23}}$) and $b'$ to either $p_{1}$ = 8-bit precision ($b' = {1}/{2^{8}}$) or $p_{1}$ = 4-bit ($b' = {1}/{2^{4}}$).
\end{itemize}
% \vspace{-7pt}
% \paragraph{None} A baseline with no compression and no multiplicative noise.
% \vspace{-7pt}
% \paragraph{L2} A magnitude pruning baseline based on the L2 Norm of weight vectors (matrices in the case of convolutional filters) at the input of each neuron to be pruned~\cite{DBLP:conf/iclr/0022KDSG17}. We set the pruning threshold to the compression rate achieved by \methodn{} using the same settings of a given experiment.
% \vspace{-7pt}

\begin{figure}[t!]
    \centering
    \begin{subfigure}[t]{0.498\textwidth}
        \centering
    \includegraphics[width=0.48\textwidth]{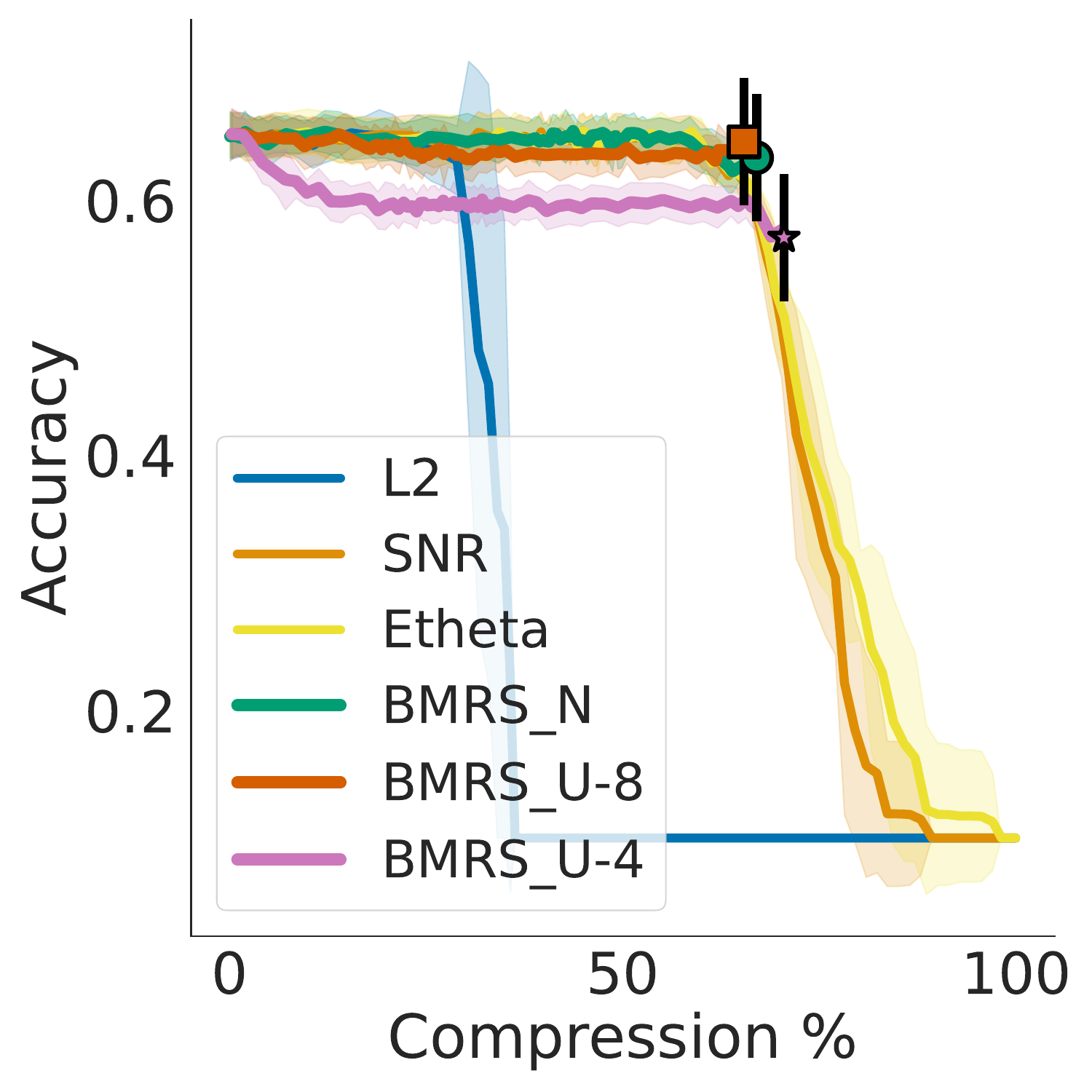}
    \includegraphics[width=0.48\textwidth]{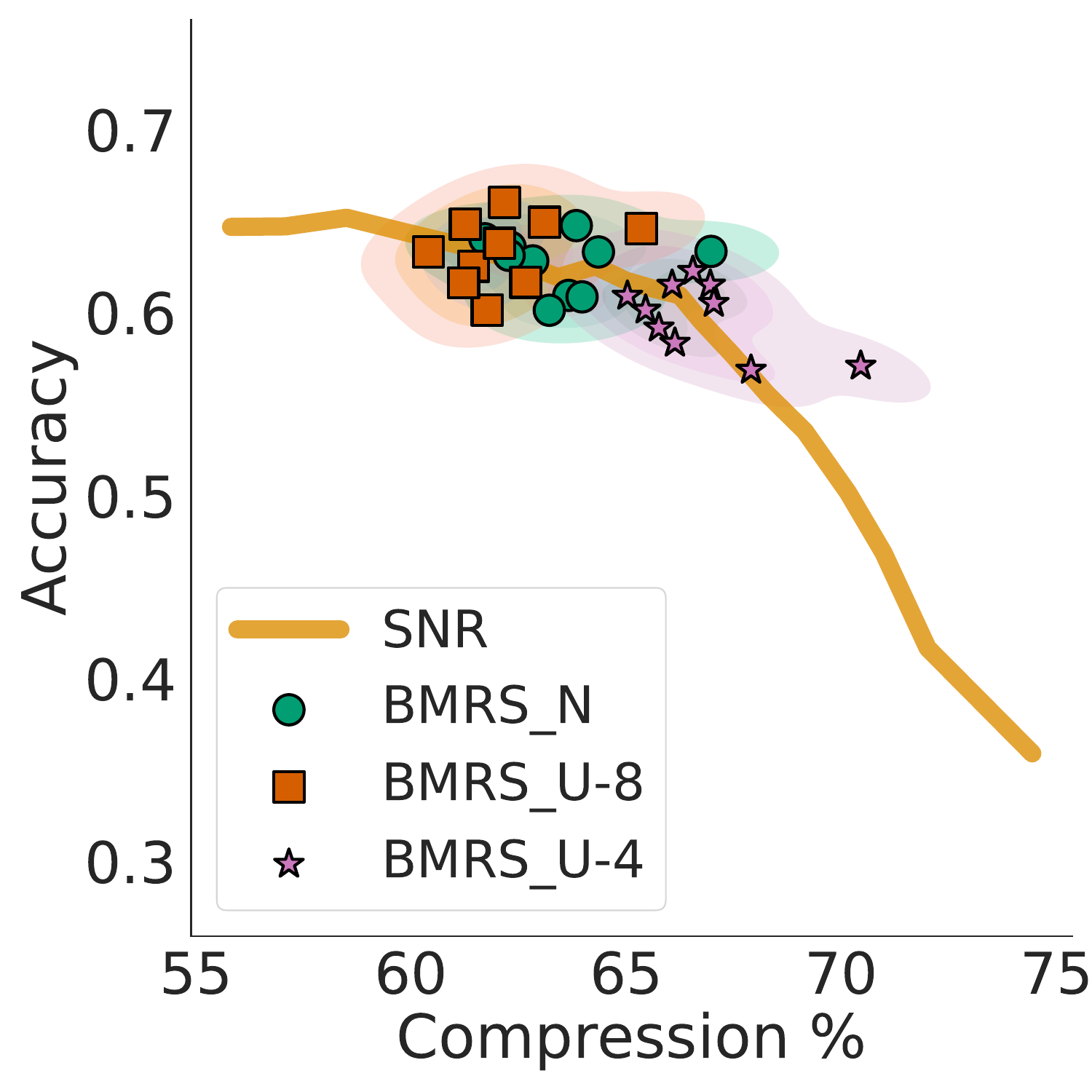}
        \caption{CIFAR10 Lenet5}
    \end{subfigure}%
    ~
    \begin{subfigure}[t]{0.498\textwidth}
        \centering
        \includegraphics[width=0.48\textwidth]{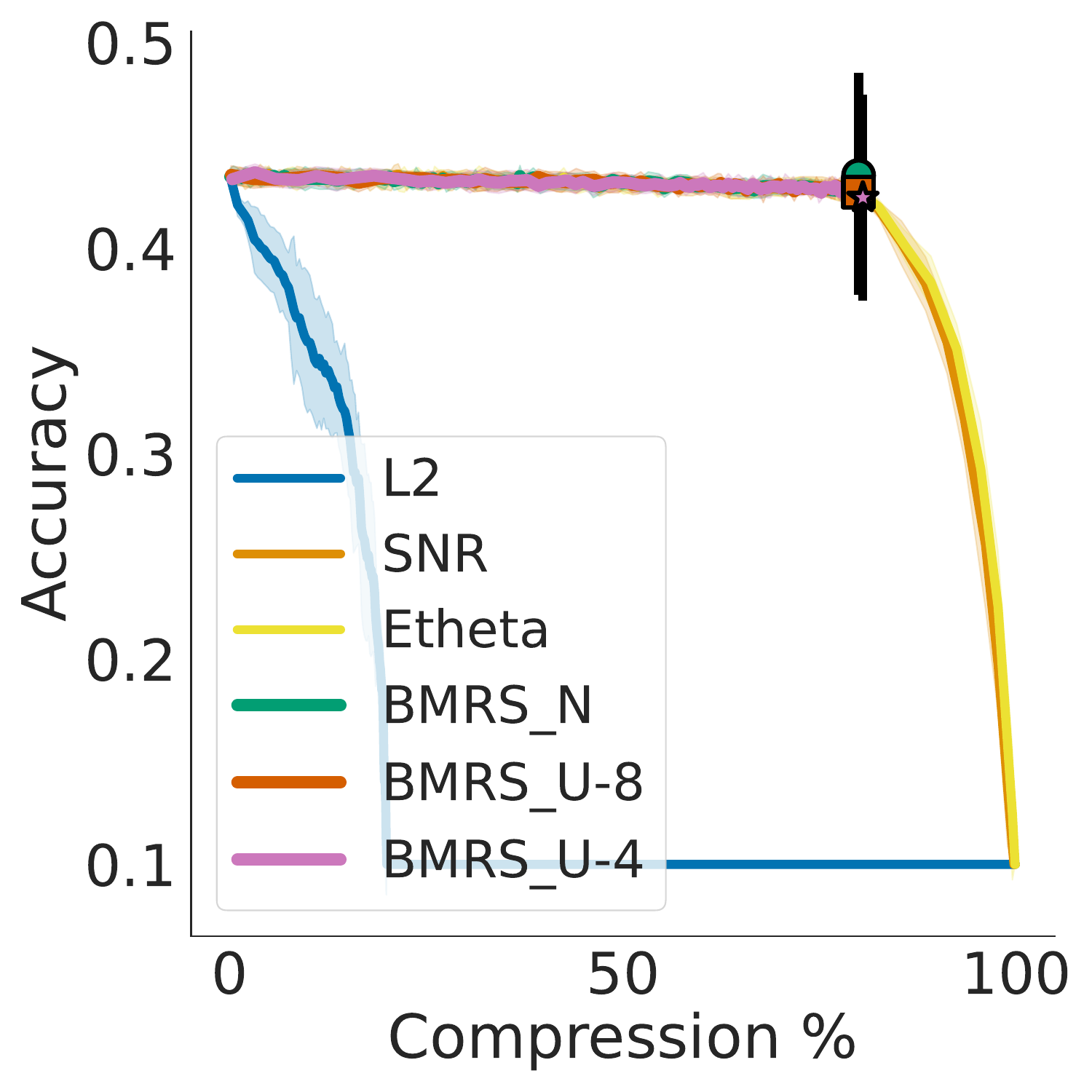}
        \includegraphics[width=0.48\textwidth]{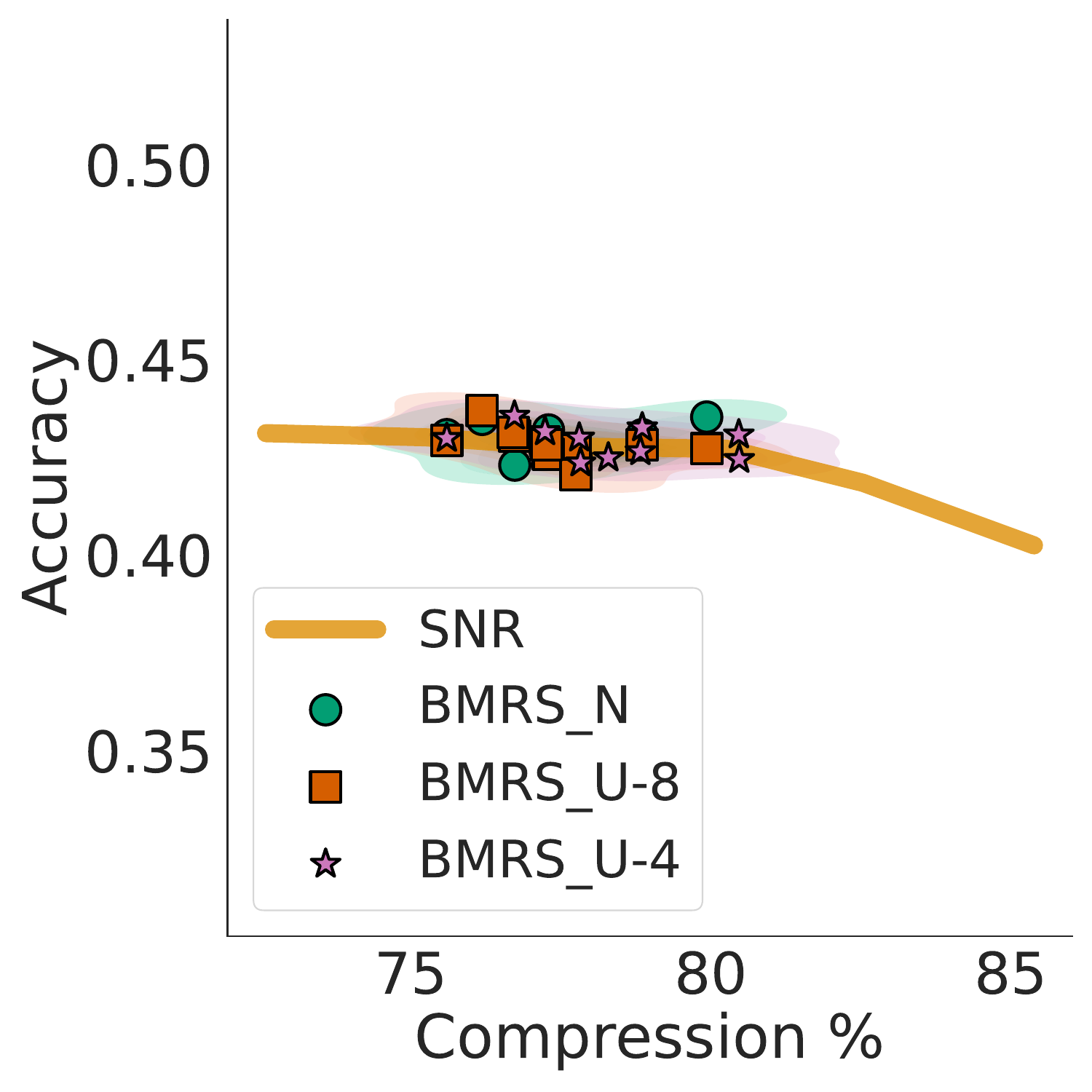}
        \caption{CIFAR10 MLP}
    \end{subfigure}% 
    \\
    \begin{subfigure}[t]{0.498\textwidth}
        \centering
    \includegraphics[width=0.48\textwidth]{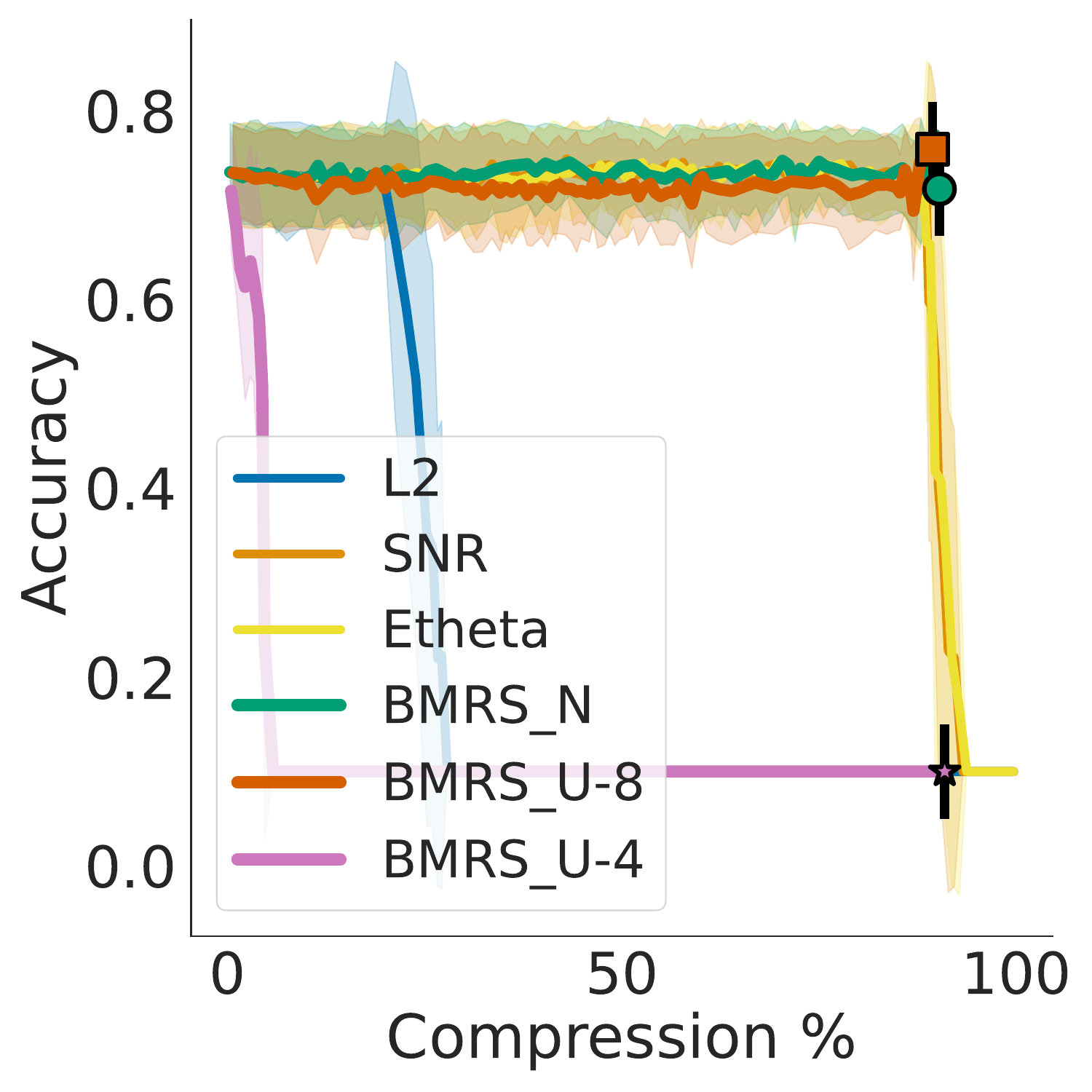}
    \includegraphics[width=0.48\textwidth]{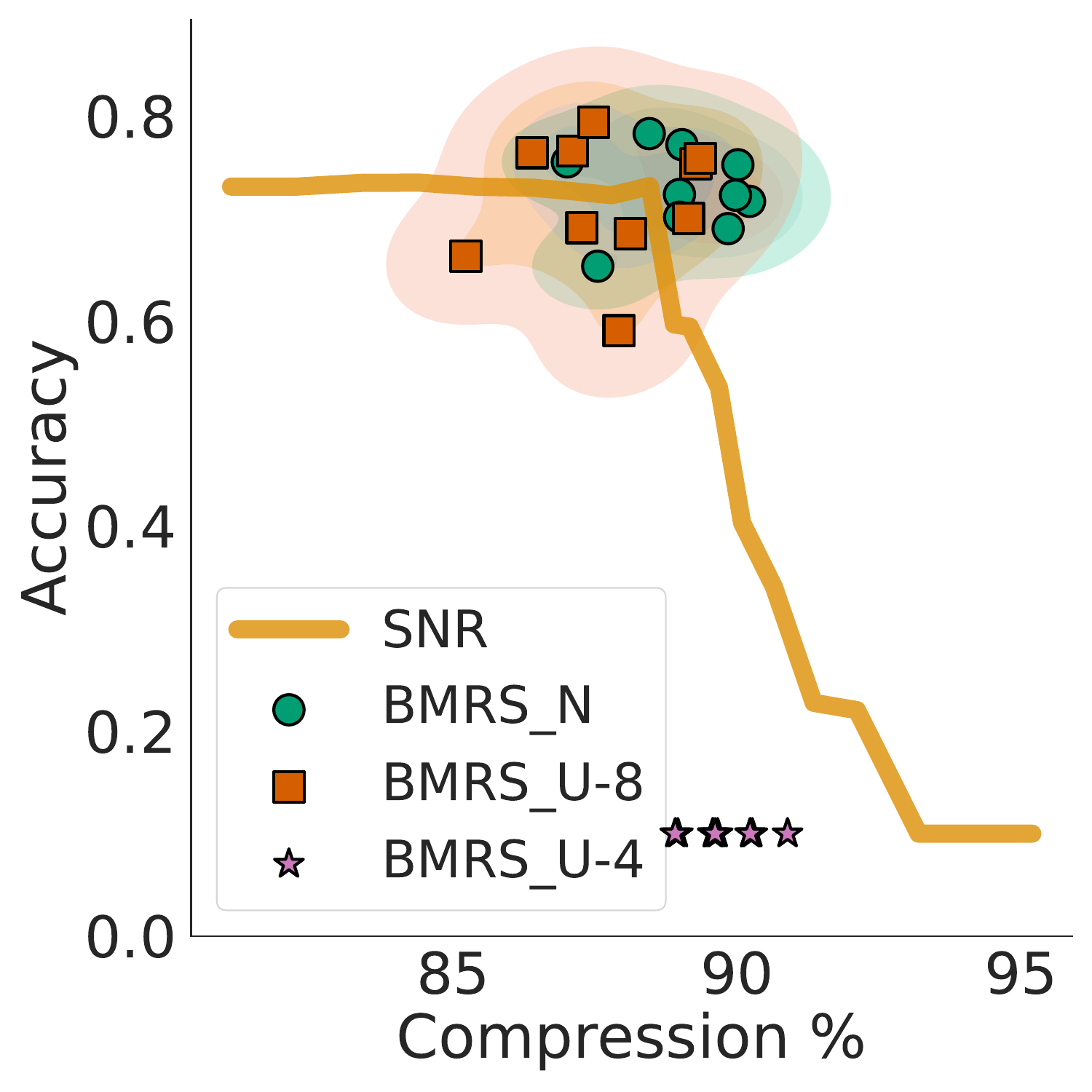}
        \caption{Fashion-MNIST Lenet5}
    \end{subfigure}%
    ~
    \begin{subfigure}[t]{0.498\textwidth}
        \centering
        \includegraphics[width=0.48\textwidth]{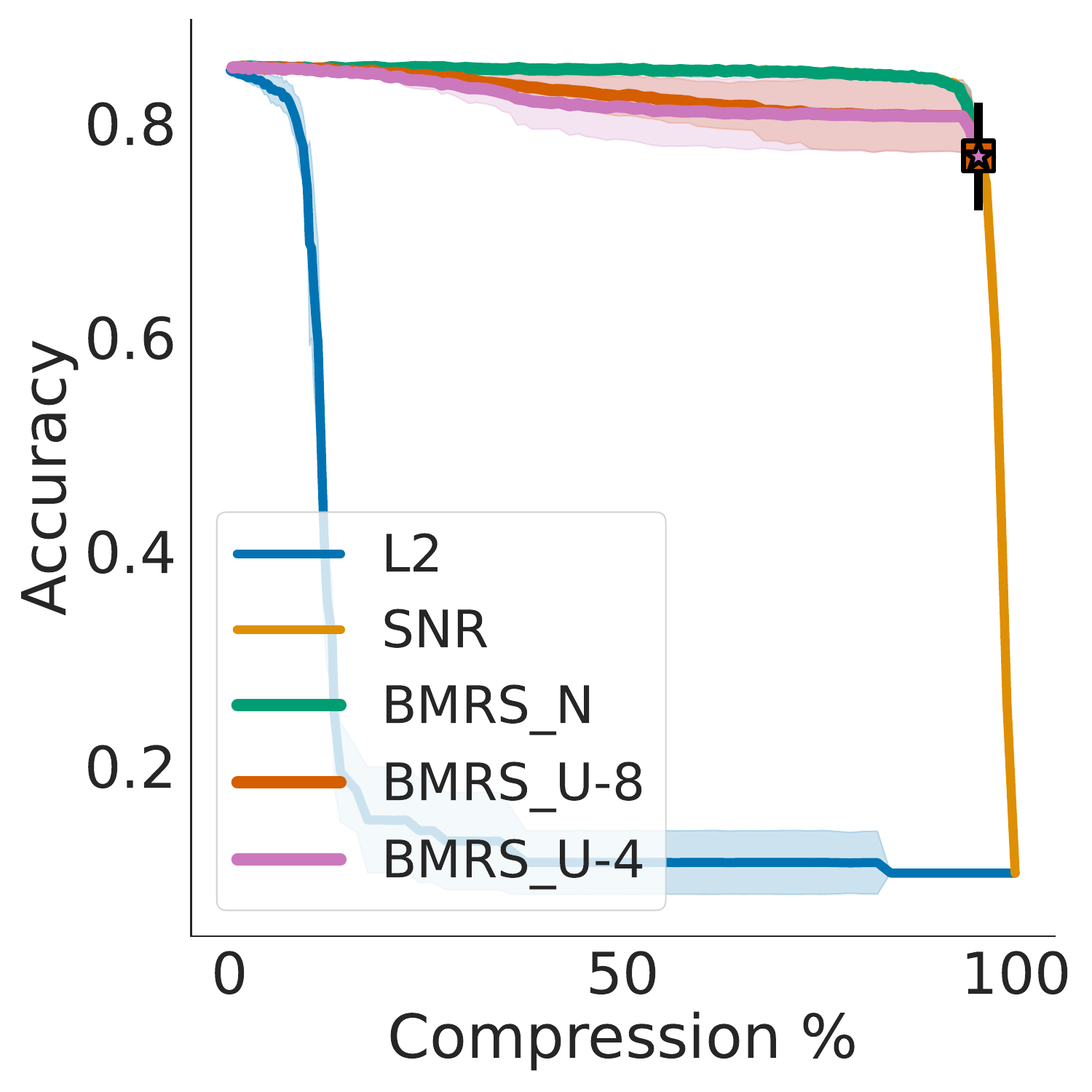}
        \includegraphics[width=0.48\textwidth]{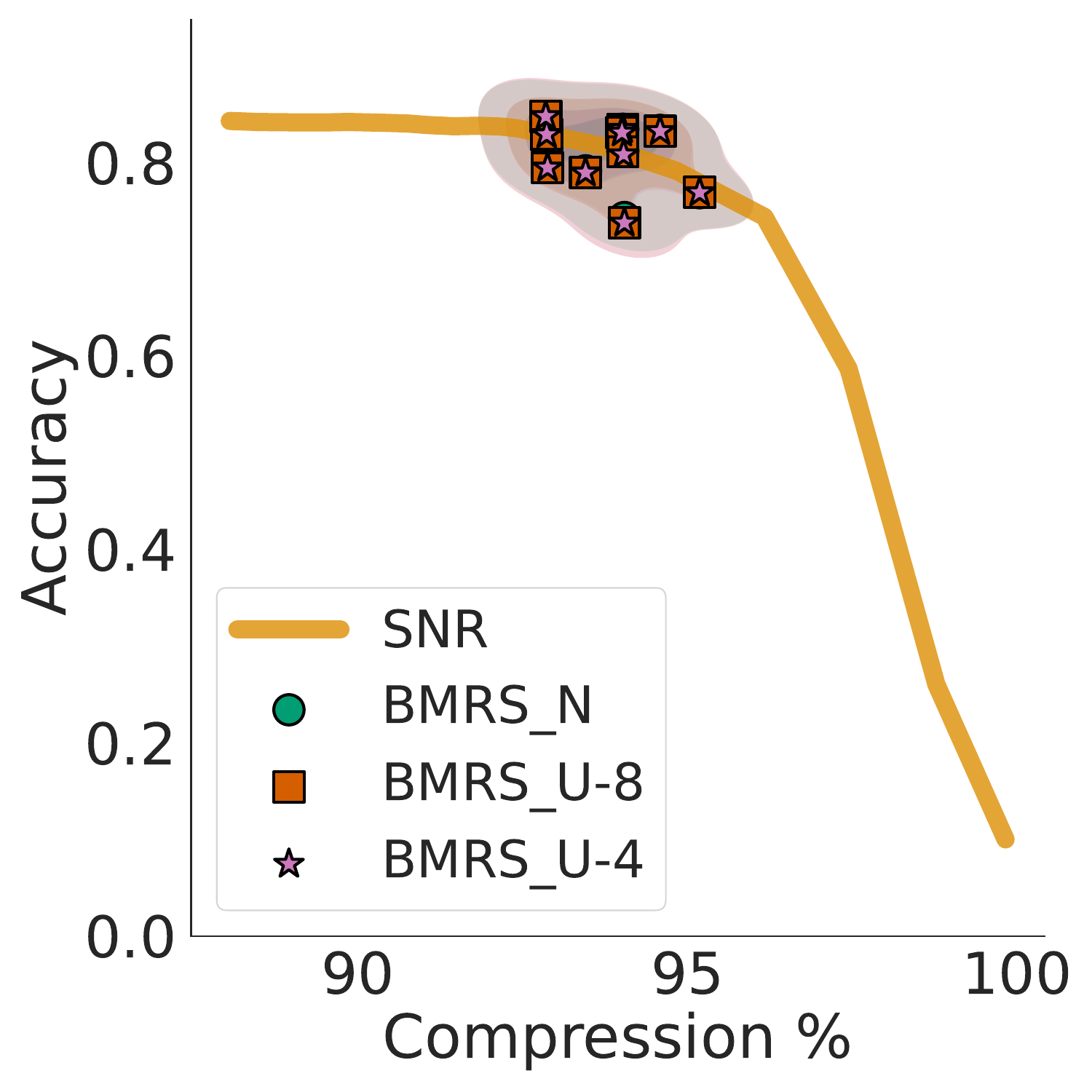}
        \caption{Fashion-MNIST MLP}
    \end{subfigure}%
    \\
    \begin{subfigure}[t]{0.498\textwidth}
        \centering
    \includegraphics[width=0.48\textwidth]{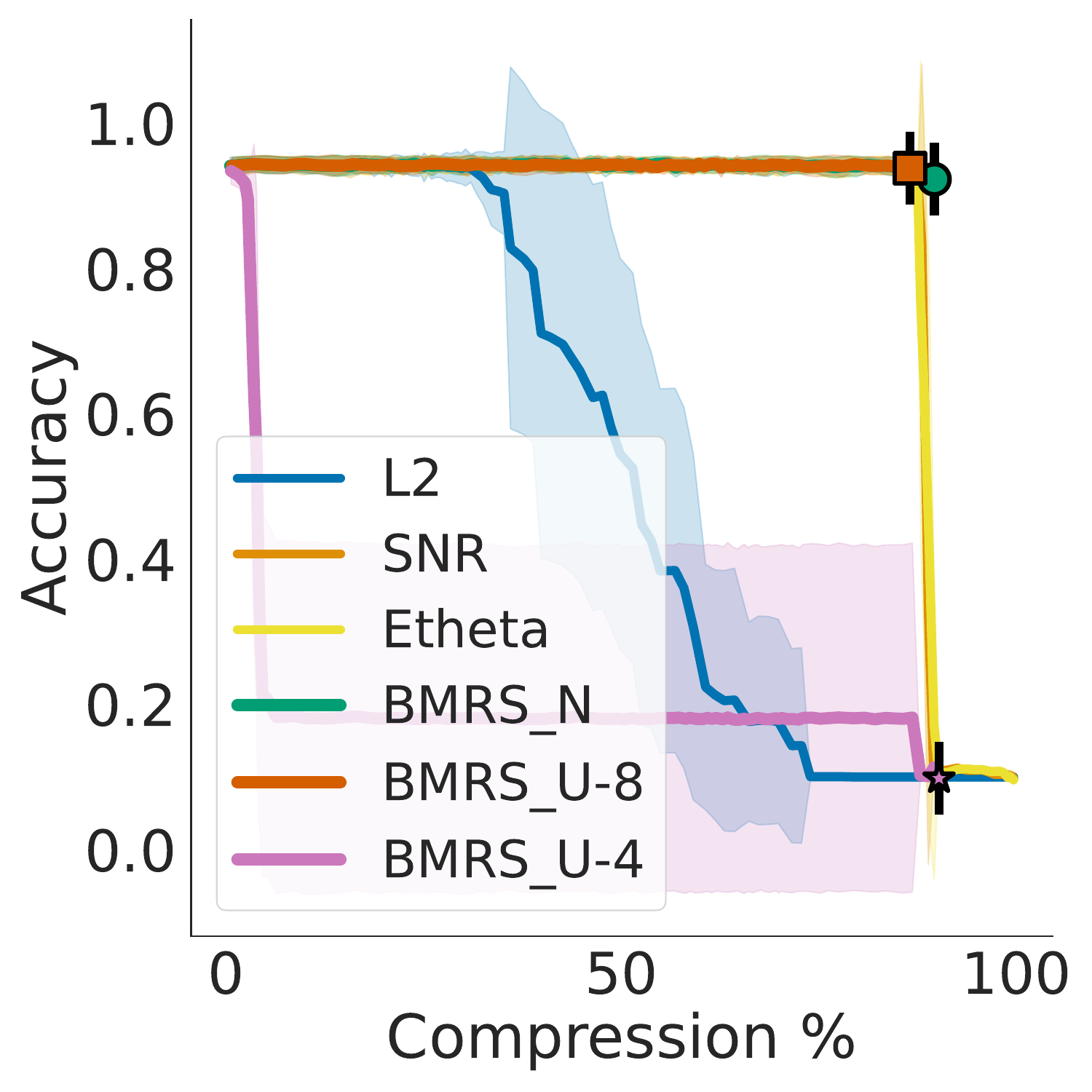}
    \includegraphics[width=0.48\textwidth]{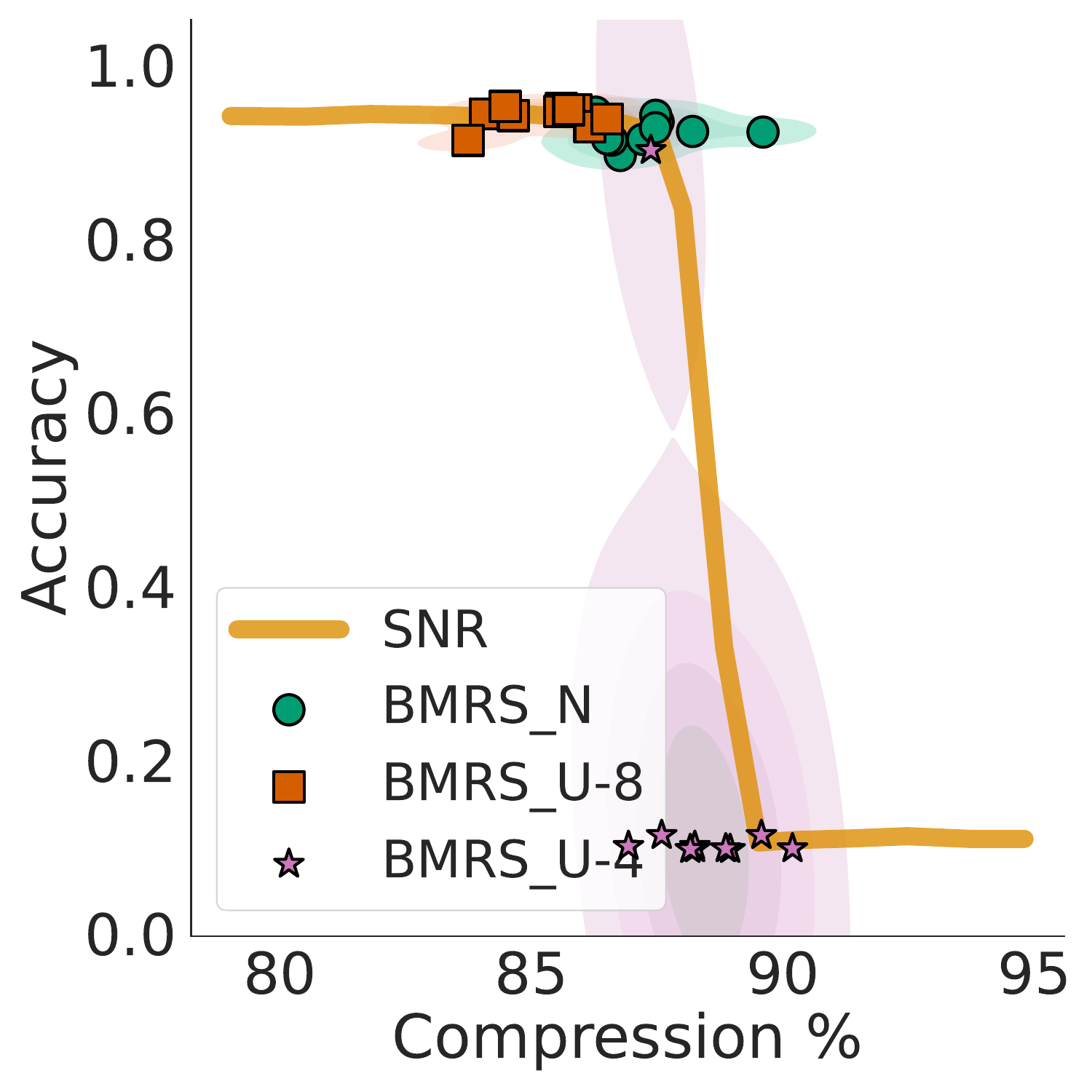}
        \caption{MNIST Lenet5}
    \end{subfigure}%
    ~
    \begin{subfigure}[t]{0.498\textwidth}
        \centering
        \includegraphics[width=0.48\textwidth]{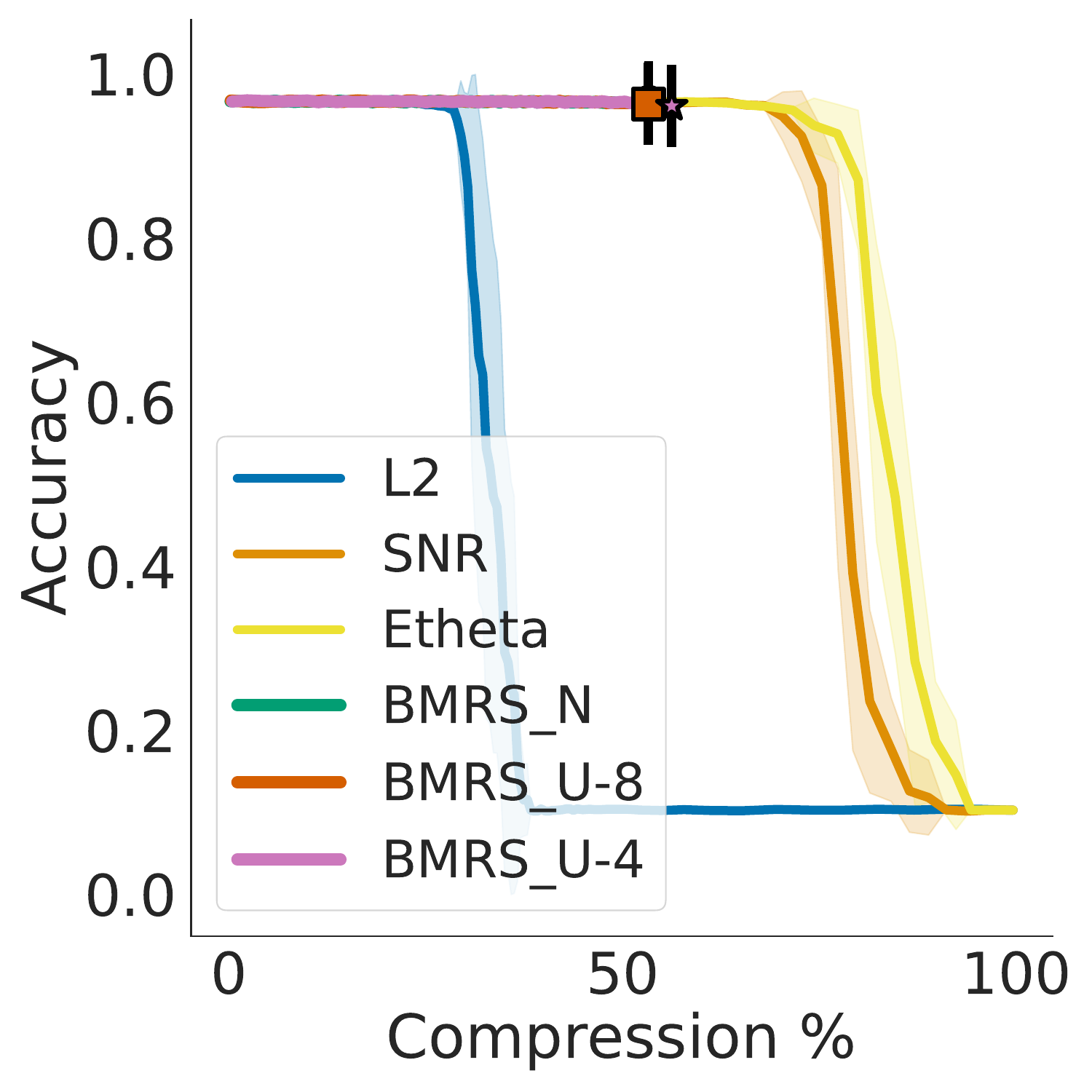}
        \includegraphics[width=0.48\textwidth]{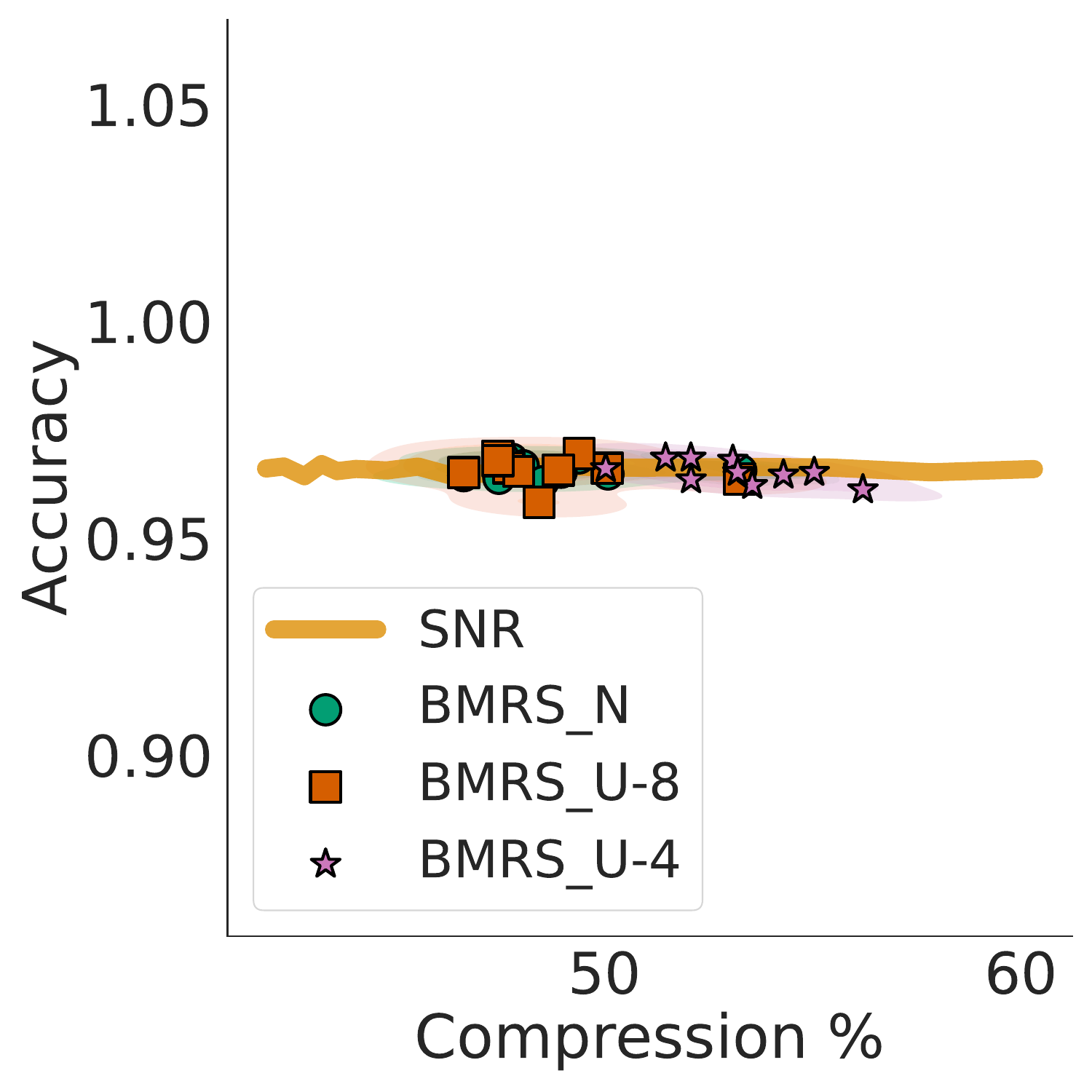}
        \caption{MNIST MLP}
    \end{subfigure}%
    \caption{Accuracy vs.~compression for post-training pruning  on CIFAR10, Fashion-MNIST, and MNIST. The left plot in each subfigure shows the average accuracy across 10 seeds, shading shows the standard deviation. For \method{}, we mark the maximum compression rate based on when $\Delta F \ge 0$.  The right plot in each subfigure shows a scatter plot and kernel density estimation of accuracy vs. compression of \method{} compared to SNR accuracy. \methodn{} and \methodu{}-8 consistently stop pruning near the knee point, a preferred trade-off solution.}
    \label{fig:post_training_graphs}

\end{figure}

%\subsection{Post-training pruning}
\paragraph{Post-training pruning.}
We first look at the behavior of \method{} when used in the post-training setting. To do so, we first train a model on a given dataset, then use each method to rank the neurons based on their pruning function (L2 norm, signal-to-noise ratio, or $\Delta F$). To observe the accuracy at different compression rates, neurons are progressively removed based on their rank, and the model is fine-tuned for one epoch before measuring the test accuracy. For \method{} methods, we additionally stop pruning once $\Delta F < 0$ for a given structure. The plots of accuracy vs. compression for 10 different random seeds are given in \autoref{fig:post_training_graphs} (Further experiments given in \autoref{sec:additional_plots}).

First, we find that \method{} generally  stops compressing near the 
%Pareto-optimal compression 
knee point of the trade-off curve -- a preferred solution of a Pareto front if there is no a priori preferences --  in all settings except for \methodu{}-4 which only does so in 4 out of 6 settings. Notably, \methodn{} accomplishes this with no need to tune additional thresholds as is common in pruning literature. To further visualize this, the right plot in each subfigure shows a scatter plot of the accuracy at the maximum compression rate (pruning all neurons where $\Delta F \ge 0$) along with the curve of accuracy vs. compression for SNR pruning near the knee point. We can see that the density of points for \method{} is concentrated near the optimal point in all cases except for the MLP on MNIST, indicating the robustness of the proposed methods.

\begin{figure}[t!]
    \centering
    \begin{minipage}[t]{.49\textwidth}
    \centering
    \begin{subfigure}[t]{0.49\textwidth}
        \centering
    \adjincludegraphics[width=1\linewidth,valign=T]{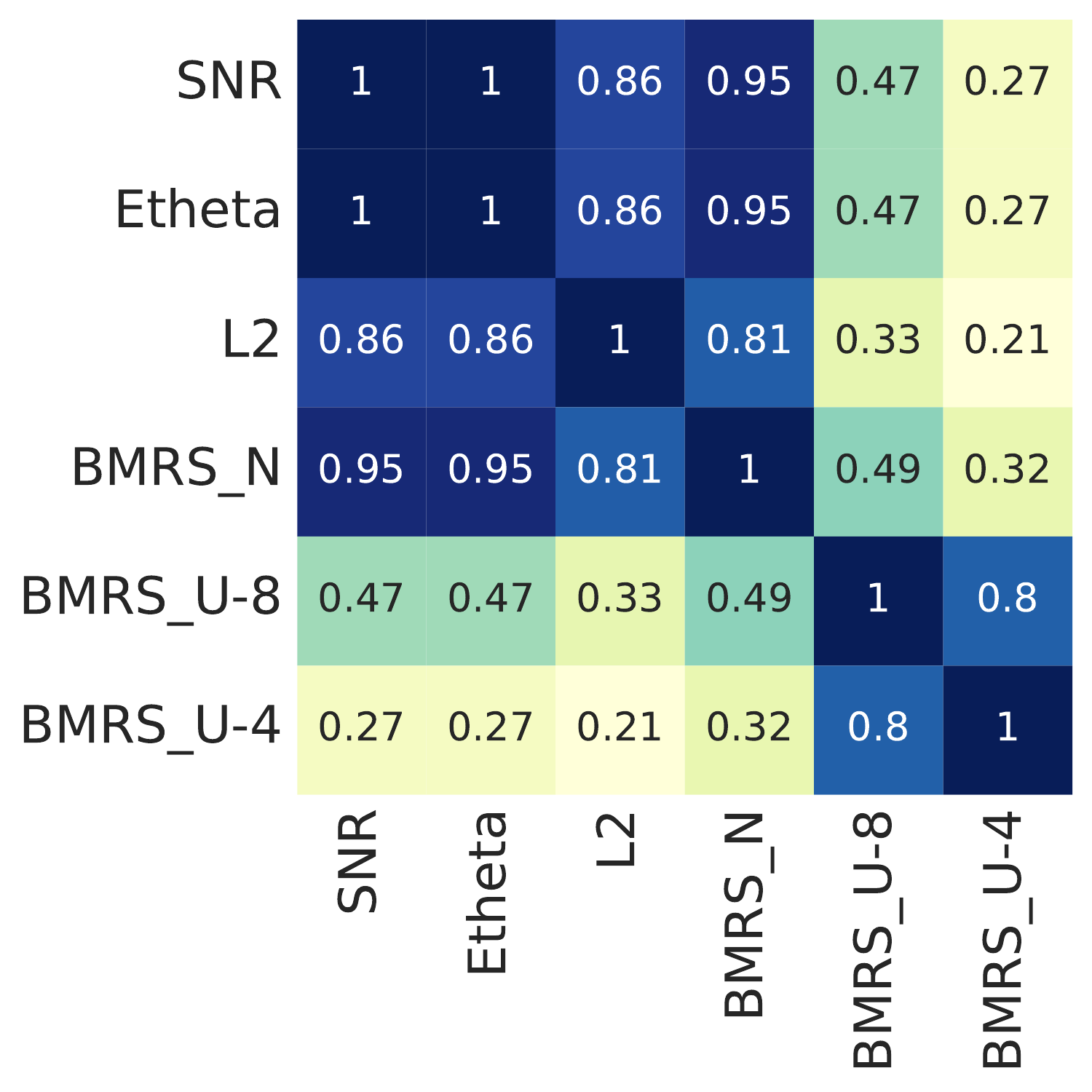}
        \caption{CIFAR10 Lenet5}
    \end{subfigure}%
    ~
    \begin{subfigure}[t]{0.49\textwidth}
        \centering
        \adjincludegraphics[width=1\linewidth,valign=T]{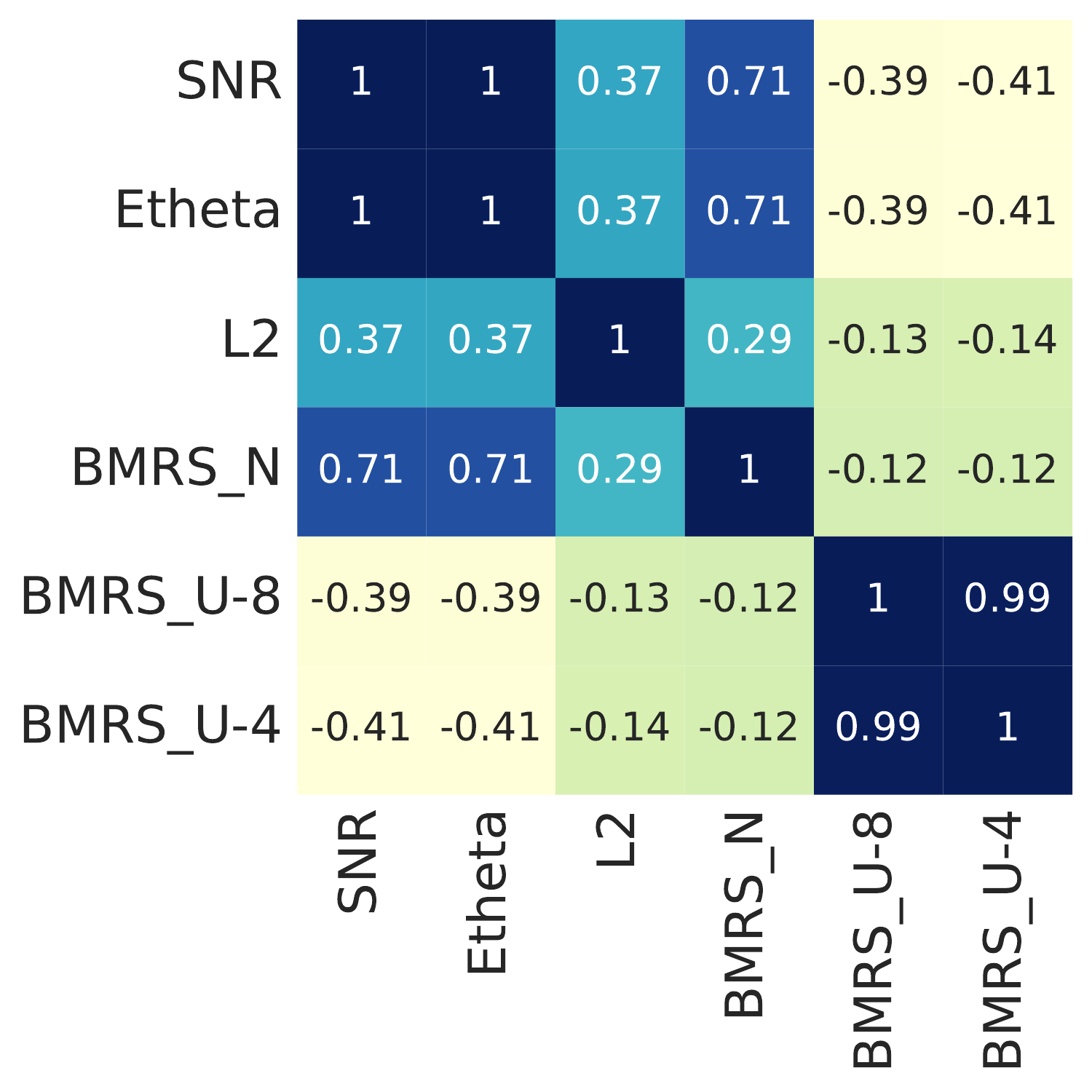}
        \caption{CIFAR10 MLP}
    \end{subfigure}%
    \caption{Average Spearman's rank correlation between the ranks of neurons for pruning when using different methods on CIFAR10 (plots for additional datasets are given in \autoref{sec:additional_plots}).} %\methodn{} tends to correlate highly with SNR, while \methodu{} can often have a negative correlation. This suggests that \methodn{} performs a qualitatively similar operation as SNR while automatically finding the compression threshold through $\Delta F$, while \methodu{} performs a qualitatively different operation.}
    \label{fig:correlations}
    \end{minipage}
    ~
    \begin{minipage}[t]{.49\textwidth}
        \centering
        \adjincludegraphics[height=3.9cm,valign=T]{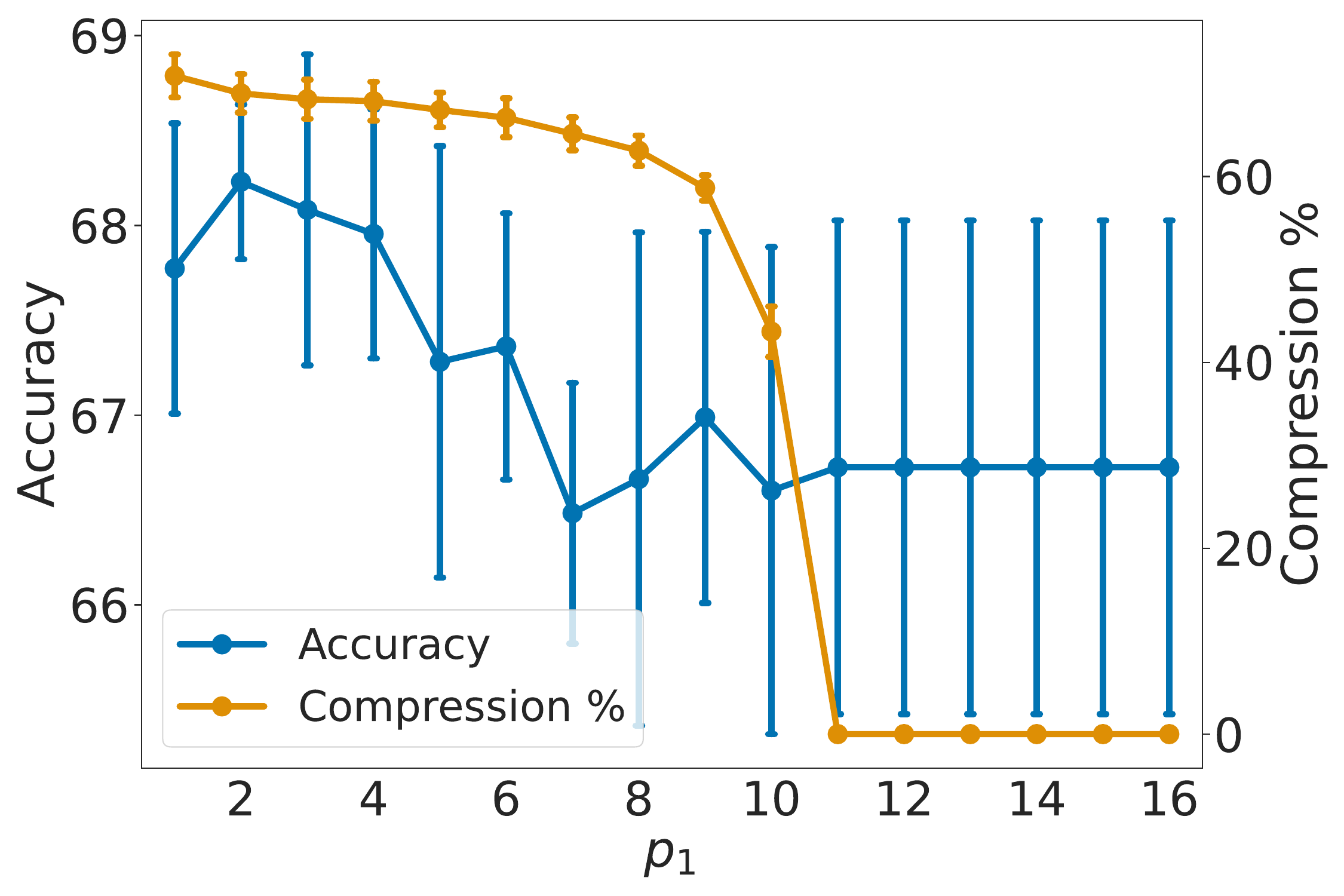}
        \caption{Accuracy and compression rate vs. $p_{1}$ for \methodu{} on CIFAR10 with Lenet5. Results are averaged across 10 seeds with standard deviation indicated by the error bars.}
        \label{fig:p1_cifar10_lenet5}
    \end{minipage}

\end{figure}

We additionally observe much similarity in the curves for \method{} and SNR pruning, suggesting that they may be performing similar functions. To further investigate this, we look at the Spearman rank correlation coefficient~\cite{sedgwick2014spearman} of the neurons based on their respective functions in \autoref{fig:correlations} (plots for additional datasets in \autoref{sec:additional_plots}). We see that \methodn{} tends to have a high correlation with SNR, suggesting that it learns a qualitatively similar function with the benefit of providing a threshold for compression. \methodu{}, on the other hand, tends to have very low or even negative correlation. This, combined with the more rapidly declining accuracy for a given level of compression, suggests that \methodu{} 
%can provide a pruning threshold but 
is only apt for determining a single split into elements to keep and to remove, but does not provide an
accurate ranking of the elements..

\paragraph{Continuous pruning.}

\begin{table}%[htp]
    \def\arraystretch{1.}
    \centering
    %\fontsize{10}{10}\selectfont
    \scriptsize
    %\rowcolors{2}{gray!10}{white}
    \caption{Parameter compression \% and accuracy for different baseline methods and settings of the proposed method. Standard deviations over ten runs are included. Best accuracy for the compression methods is given in bold.} %Results are reported as the average macro F1 over 5 random seeds.}
    \label{tab:small_results}
    \begin{tabular}{l c c c c  c c}
    \toprule %\thickhline
    & \multicolumn{2}{c}{\bf MNIST } & \multicolumn{2}{c}{\bf Fash-MNIST } & \multicolumn{2}{c}{\bf CIFAR10 } \\
    \cmidrule{2-3}
    \cmidrule{4-5}
    \cmidrule{6-7}
    {\bf Pruning Method} & {\bf Comp. (\%)} & {\bf Acc.} & {\bf Comp. (\%)} &{\bf Acc.} & {\bf Comp. (\%)} & {\bf Acc.}\\

\cmidrule{2-7}
& \multicolumn{6}{c}{\bf MLP}\\
\cmidrule{2-7}
None&$0.00 \pm 0.00$& $97.43 \pm 0.14$& $0.00 \pm 0.00$& $88.17 \pm 0.20$& $0.00 \pm 0.00$& $44.94 \pm 0.40$\\
L2&$43.11 \pm 2.06$& $10.39 \pm 0.32$& $87.86 \pm 2.27$& $18.23 \pm 10.22$& $42.89 \pm 2.64$& $10.00 \pm 0.00$\\
$E[\theta]$&$52.08 \pm 1.71$& $96.88 \pm 0.15$& $91.76 \pm 0.81$& $85.59 \pm 0.26$& $77.99 \pm 1.54$& $43.39 \pm 0.46$\\
SNR&$58.57 \pm 2.01$& $96.92 \pm 0.08$& $99.83 \pm 0.00$& $10.00 \pm 0.00$& $75.93 \pm 1.26$& $43.97 \pm 0.46$\\
\methodn{}&$48.86 \pm 1.32$& $96.95 \pm 0.19$& $93.20 \pm 0.66$& $84.99 \pm 0.35$& $76.36 \pm 1.08$& $43.59 \pm 0.29$\\
\methodu{}-8&$48.73 \pm 1.90$& $96.93 \pm 0.16$& $93.02 \pm 0.81$& $85.01 \pm 0.32$& $77.17 \pm 0.98$& $43.45 \pm 0.42$\\
\methodu{}-4&$54.47 \pm 1.74$& $\mathbf{96.99 \pm 0.13}$& $91.57 \pm 0.71$& $\mathbf{85.79 \pm 0.34}$& $76.63 \pm 0.94$& $\mathbf{44.06 \pm 0.40}$\\
\cmidrule{2-7}
& \multicolumn{6}{c}{\bf Lenet5}\\
\cmidrule{2-7}
None&$0.00 \pm 0.00$& $99.07 \pm 0.09$& $0.00 \pm 0.00$& $89.16 \pm 0.27$& $0.00 \pm 0.00$& $67.62 \pm 0.77$\\
L2&$83.42 \pm 1.92$& $11.35 \pm 0.00$& $83.62 \pm 1.69$& $10.00 \pm 0.00$& $52.29 \pm 2.18$& $10.00 \pm 0.00$\\
$E[\theta]$&$88.29 \pm 1.00$& $51.30 \pm 41.12$& $89.71 \pm 0.56$& $50.93 \pm 33.45$& $66.19 \pm 1.36$& $65.83 \pm 0.90$\\
SNR&$92.66 \pm 5.77$& $62.70 \pm 41.93$& $98.47 \pm 3.45$& $17.01 \pm 21.03$& $70.29 \pm 2.02$& $\mathbf{67.68 \pm 0.52}$\\
\methodn{}&$86.90 \pm 1.15$& $95.59 \pm 0.94$& $88.02 \pm 1.00$& $77.90 \pm 2.44$& $62.87 \pm 1.64$& $66.14 \pm 0.70$\\
\methodu{}-8&$86.11 \pm 1.37$& $95.27 \pm 1.02$& $87.61 \pm 0.72$& $77.23 \pm 3.49$& $62.54 \pm 1.49$& $66.28 \pm 1.07$\\
\methodu{}-4&$87.58 \pm 1.01$& $\mathbf{96.66 \pm 0.59}$& $88.72 \pm 0.73$& $\mathbf{81.10 \pm 1.50}$& $68.07 \pm 1.95$& $67.66 \pm 0.59$\\

    \bottomrule % \thickhline

    \end{tabular}
    
\vspace{-10pt}
\end{table}

Next, we experiment with continuous pruning, where neurons are pruned continuously throughout training based on either a provided pruning threshold (SNR, ${\mathbb{E}_{q_{\mathbf{\phi}}}[\theta]}$) or $\Delta F$ (\method{}). For SNR, we prune a neuron when its SNR falls below 1 (as in \cite{DBLP:conf/nips/NeklyudovMAV17}), and for ${\mathbb{E}_{q_{\mathbf{\phi}}}[\theta]}$ we set a threshold of 0.1. For L2 pruning, we perform post-training pruning based on the compression rate achieved by \methodn{}. Neurons are pruned after every epoch during training, followed by 10 epochs of fine-tuning at the very end of training.

We compare the raw performance of each variant using an MLP and Lenet5 on MNIST, Fashion-MNIST, and CIFAR10 in \autoref{tab:small_results}. First, we note that using the L2 norm with the same compression rate as \methodn{} results in a degenerate model; the accuracy degrades to random in all settings. Additionally, we see that using the SNR as a pruning criterion with the recommended threshold of 1 from \cite{DBLP:conf/nips/NeklyudovMAV17} is also inconsistent, resulting in large drops in performance for 3 out of 6 settings. \methodn{} and \methodu{} result in both high compression rate and high performance in all settings. \methodn{} accomplishes this without the need for tuning any pruning thresholds, in one case yielding a higher compression rate than \methodu{}-4 while keeping the accuracy high. \methodu{}-4 results in both the highest compression rate among the three \method{} variants in 4 out of 6 settings, and the highest accuracy in 5 out of 6 settings, with the caveat of needing to select $p_{1}$ as a hyperparameter. To explore the effect of this hyperparameter further, we plot accuracy and compression vs. $p_{1}$ for Lenet5 trained on CIFAR10 in \autoref{fig:p1_cifar10_lenet5}. We see that compression rapidly increases after $p_{1} = 11$, continuing until $p_{1} = 1$. Additionally, we find that accuracy also steadily increases with a higher compression rate, indicating that \methodu{} reduces complexity while increasing the generalization capacity of the model. 

Finally, a comparison of ResNet-50 and ViT on CIFAR10 and TinyImagenet is given in \autoref{tab:large_results}. Here, SNR and the three \method{} variants achieve similar accuracies at different compression rates. \methodn{} achieves a modest compression rate compared to SNR with a threshold of 1 for each case. \methodu{}-4 yields a higher compression rate than \methodn{} and \methodu{}-8 in all settings. As such, we show that \methodn{} is capable of achieving high compression with no threshold tuning, while a more extreme compression rate is possible by selecting $p_{1}$ for \methodu{}. 

\begin{table}%[htp]
    \def\arraystretch{1.2}
    \centering
    %\fontsize{10}{10}\selectfont
    \scriptsize
    %\rowcolors{2}{gray!10}{white}
    \caption{Parameter compression \% and accuracy for different baseline methods and settings of the proposed method. Standard deviations over three runs are included. Best accuracy for the compression methods is given in bold.} 
    % Standard deviation given in subscript.} %Results are reported as the average macro F1 over 5 random seeds.}
    \begin{tabular}{l c c  c c}
    \toprule %\thickhline
    & \multicolumn{2}{c}{\bf CIFAR10 } & \multicolumn{2}{c}{\bf TinyImagenet } \\
    \cmidrule{2-5}
    {\bf Pruning Method} & {\bf Comp. \%} & {\bf Acc.} & {\bf Comp. \%} & {\bf Acc.}\\

\cmidrule{2-5}
& \multicolumn{4}{c}{\bf Res50-Pretrained}\\
\cmidrule{2-5}
None&$0.00 \pm 0.00$& $90.65 \pm 0.05$& $0.00 \pm 0.00$& $53.01 \pm 0.35$\\
L2&$63.79 \pm 4.21$& $10.00 \pm 0.00$& $55.71 \pm 1.08$& $0.50 \pm 0.00$\\
$E[\theta]$&$89.85 \pm 0.09$& $16.27 \pm 1.86$& $83.59 \pm 1.50$& $21.11 \pm 6.47$\\
SNR&$91.73 \pm 0.16$& $89.24 \pm 0.40$& $77.85 \pm 0.14$& $50.54 \pm 0.59$\\
\methodn{}&$87.10 \pm 1.55$& $\mathbf{89.62 \pm 0.41}$& $74.98 \pm 0.16$& $50.56 \pm 0.33$\\
\methodu{}-8&$88.18 \pm 0.22$& $89.29 \pm 0.20$& $74.99 \pm 0.04$& $\mathbf{50.84 \pm 0.34}$\\
\methodu{}-4&$89.85 \pm 0.05$& $89.26 \pm 0.28$& $76.12 \pm 0.13$& $50.82 \pm 0.50$\\
\cmidrule{2-5}
& \multicolumn{4}{c}{\bf Vision Transformer}\\
\cmidrule{2-5}
None&$0.00 \pm 0.00$& $94.80 \pm 0.17$& $0.00 \pm 0.00$& $63.14 \pm 0.42$\\
L2&$57.74 \pm 0.26$& $54.36 \pm 1.84$& $47.47 \pm 0.12$& $7.08 \pm 0.57$\\
$E[\theta]$&$68.18 \pm 0.09$& $10.00 \pm 0.00$& $57.08 \pm 0.18$& $0.50 \pm 0.00$\\
SNR&$73.03 \pm 0.03$& $94.78 \pm 0.10$& $62.75 \pm 0.04$& $64.60 \pm 0.07$\\
\methodn{}&$57.74 \pm 0.25$& $94.60 \pm 0.01$& $47.48 \pm 0.12$& $65.00 \pm 0.21$\\
\methodu{}-8&$58.14 \pm 0.11$& $94.69 \pm 0.04$& $47.34 \pm 0.14$& $\mathbf{65.14 \pm 0.13}$\\
\methodu{}-4&$67.16 \pm 0.21$& $\mathbf{94.83 \pm 0.25}$& $56.13 \pm 0.14$& $65.13 \pm 0.10$\\

    \bottomrule % \thickhline

    \end{tabular}
    \label{tab:large_results}
\end{table}

% \begin{figure}
%     \centering
%     \includegraphics[width=0.48\textwidth]{figures/cifar10_lenet5_bmrsu.pdf}
%     \caption{Accuracy and compression rate vs. $p_{1}$ for \methodu{} on CIFAR10 with Lenet5. Results are averaged across 3 seeds with 95\% confidence intervals indicated by the error bars.}
%     \label{fig:p1_cifar10_lenet5}
% \end{figure}
%\paragraph{Effect of $p_{1}$} We briefly highlight the effect of changing $p_{1}$ on accuracy and compression rate for \methodu{} in \autoref{fig:p1_cifar10_lenet5} by plotting both accuracy and compression vs. $p_{1}$ for Lenet5 trained on CIFAR10. We see that the compression rate rapidly increases at $p_{1} = 8$, then steadily increases until $p_{1} = 2$. Additionally, we find that accuracy also steadily increases with a higher compression rate, indicating that \methodu{} reduces complexity while increasing the generalization capacity of the model. 

\section{Discussion and conclusion}
\label{sec:conclusion}

% \begin{enumerate}
%     \item Summarize and compare use of BMRS in the post-training and continuous pruning settings. Does the method lend itself better suited for one of these settings? What would we recommend, and why?
%     \item Discuss how the bit length for \methodu{} can be used for future applications; can we provide some more insight into when which length we should use? Or at least mention this can be tuned as a hyperparameter
%     \item A general discussion on the choice of priors? Can we say something about when to use \methodn{} and \methodu{}? Based on our experiments but more forward looking for other applications. 
% \end{enumerate}

Our experimental results demonstrate the pruning characteristics of \method{} in two settings: continuous pruning and post-training pruning. One of the benefits of \method{}, and \methodn{} in particular, is that no threshold tuning is needed, making it particularly useful for the continuous pruning setting. Here, the compression rate improves as the model converges, allowing one to gradually increase the compression rate without sacrificing accuracy. The choice of \methodn{} vs. \methodu{} is then dependent on the problem, where more complex scenarios (e.g., over-parameterized models) are suitable for \methodu{} if a higher compression rate is desired. In this case, a hyperparameter search over $p_{1}$ in the range $[\epsilon, 23]$ can be done to select \methodu{} (see \autoref{fig:p1_cifar10_lenet5} for an example), or one can simply use \methodn{} to achieve good compression with no hyperparameter tuning. Additionally, we expect that the more over-specified the model is, the lower we can set $p_{1}$ and maintain accuracy (i.e., lower bit-rate).

{\bf Limitations:} We note a few of the limitations of \method{}. First, while multiplicative noise pruning allows for the flexible application of pruning at different structural levels, BNNs may offer more aggressive compression rates as they apply sparsity inducing priors at multiple hierarchical levels~\cite{DBLP:conf/nips/BaiSC20,DBLP:journals/corr/abs-2305-00934,DBLP:journals/nn/JantreBM23,sun2022learning,DBLP:journals/corr/Markovic}. BMR based approaches may be derived for such networks; as of this work and as far we we know, it has only been successfully applied in practice to models with flat priors for unstructured pruning~\cite{beckers2024principled,DBLP:journals/corr/Markovic}. Additionally, multiplicative noise creates an overhead of additional parameters $\mathbf{\phi}$ which increase the training time and storage requirements. Third, more exhaustive baseline pruning criteria could be used for comparison in the future, for example classic methods based on the Hessian or gradient of weights (e.g. see \autoref{fig:post_training_graphs_supplement} in \autoref{sec:additional_plots})~\cite{DBLP:conf/nips/CunDS89,DBLP:conf/nips/HassibiS92}. Fourth, we apply multiplicative noise to linear layers and convolutional filters, while it could be useful to explore pruning more complex structures in the future. Finally, while structured pruning can reduce the inference time and energy consumption of neural networks, improvements in efficiency have been shown to have potential negative consequences in terms of energy consumption and carbon emissions based on how efficiency can affect how a model is used in practice~\cite{DBLP:journals/corr/abs-2309-02065}. 

{\bf Conclusions:} In this work we presented \method{}, an efficiently calculable method for threshold-free structured pruning of neural networks. We derived two versions of \method{}: \methodn{} based on the truncated log-normal prior, and \methodu{} based on a reduced truncated log-uniform prior. \method{} offers several key features over existing work: by basing the method off of the approach of multiplicative noise~\cite{DBLP:conf/nips/NeklyudovMAV17}, the structured pruning aspect is flexible as it is not dependent on assuming any prior over individual weights and can be easily applied at any structural level~\cite{DBLP:journals/nn/JantreBM23,DBLP:conf/nips/BaiSC20}. 
%Additionally, since we use a flat prior we are able to achieve principled structured pruning with low computational complexity using BMR, which is not achieved when using hierarchical structured priors on BNNs~\cite{DBLP:journals/corr/Markovic}. 
Additionally, the prior and variational posterior in the multiplicative noise approach lend themselves to the derivation of \method{} using multiple reduced priors which have different pruning properties, allowing for flexibility in the compression rate when desired and threshold free pruning otherwise. Finally, our experimental results demonstrate the competitive compression and accuracy of \method{} compared to baseline compression methods on multiple networks of varying complexity and across multiple datasets. The methods presented here based on BMR could form a template for developing more aggressive pruning schemes by incorporating more complex hierarchical priors on both structures and individual weights, as well as for studying limits on the number of neural network structures required to solve a given dataset.

{\bf Acknowledgments} DW, CI and RS are partly funded by European Union’s Horizon Europe Research and Innovation Programme under grant agreements No. 101070284 and No. 101070408. DW is also partly supported by a Danish Data
Science Academy postdoctoral fellowship (grant:
2023-1425).

\clearpage
\bibliography{main}
\bibliographystyle{abbrvnat}
%%%%%%%%%%%%%%%%%%%%%%%%%%%%%%%%%%%%%%%%%%%%%%%%%%%%%%%%%%%%

\newpage

\appendix
\section{Derivations}
\label{sec:all_derivations}
We use the following notation and distributions: 
\begin{equation*}
\begin{split}
    &\Phi(x) = \frac{1}{2}\left[1 + \text{erf}\left(\frac{x}{\sqrt{2}}\right)\right] \quad \text{(CDF of $\mathcal{N}(0,1)$ evaluated at $x$)} \\
    &\alpha_{p} = \frac{a - \mu_{p}}{\sigma_{p}} \\
    &\beta_{p} = \frac{b - \mu_{p}}{\sigma_{p}} \\
    &Z_{p} = \Phi(\beta_{p}) - \Phi(\alpha_{p}) \\
    &\text{LogN}_{[a,b]}(\theta|\mu_{p}, \sigma_{p}^{2}) = \begin{cases}
        \frac{1}{Z_{p}\theta\sqrt{2\pi\sigma_{p}^{2}}}\exp\left\{-\frac{1}{2}\frac{(\log\theta - \mu_{p})^{2}}{\sigma_{p}^{2}}\right\} & a \le \theta \le b \\
    0, & \text{otherwise}
    \end{cases}\\
    &
    \text{LogU}_{[a, b]}(\theta) =  \begin{cases}
    \left({\theta\log \frac{b}{a}}\right)^{-1}, & a \le \theta \le b \\
    0, & \text{otherwise}
  \end{cases}
\\
    &q_{\mathbf{\phi}}(\theta) = \text{LogN}_{[a,b]}(\theta|\mu_{q}, \sigma_{q}^{2}) \\
    &\tilde{p}(\theta) = \text{LogN}_{[a,b]}(\theta|\tilde{\mu}_{p}, \tilde{\sigma}_{p}^{2}) \quad \\
    &p(\theta) = \text{LogU}_{[a,b]}(\theta)
\end{split}
\end{equation*}

\subsection{\methodn{}}
\label{sec:BMR_deriv}
We start by finding $q_{\mathbf{\phi}}(\theta)\frac{\tilde{p}(\theta)}{p(\theta)}$.

\newcommand{\three}{\left(\frac{(\log\theta - \mu_{q})^{2}}{\sigma_{q}^{2}} + \frac{(\log\theta - \tilde{\mu}_{p})^{2}}{\tilde{\sigma}_{p}^{2}}\right)}
\newcommand{\threeTwo}{\left(\frac{\mu_{q}^{2}}{\sigma_{q}^{2}} + \frac{\tilde{\mu}_{p}^{2}}{\tilde{\sigma}_{p}^{2}} - \frac{\tilde{\mu}_{q}^{2}}{\tilde{\sigma}_{q}^{2}}\right)}

1.  First we look at
\begin{equation*}
q_{\mathbf{\phi}}(\theta)\tilde{p}(\theta) = \frac{1}{\theta^{2}2\pi Z_{q}\tilde{Z}_{p}\sqrt{\sigma_{q}^{2}\tilde{\sigma}_{p}^{2}}}\exp\left\{-\frac{1}{2}\three\right\}.
\end{equation*}
The exponent can be rewritten as 
\begin{multline*}
-\frac{1}{2}\three =\\
-\frac{1}{2} \frac{\tilde{\sigma}_{p}^{2}(\log^{2}\theta - 2\log\theta\mu_{q} + \mu_{q}^{2}) + \sigma_{q}^{2}(\log^{2}\theta - 2\log\theta\tilde{\mu}_{p} + \tilde{\mu}_{p}^{2})}{\sigma_{q}^{2}\tilde{\sigma}_{p}^{2}} =\\
-\frac{1}{2} \frac{(\sigma_{q}^{2} + \tilde{\sigma}_{p}^{2})\left(\log\theta - \frac{\sigma_{q}^{2}\tilde{\sigma}_{p}^{2}}{\sigma_{q}^{2} + \tilde{\sigma}_{p}^{2}}\left(\frac{\mu_{q}}{\sigma_{q}^{2}} + \frac{\tilde{\mu_{p}}}{\tilde{\sigma}_{p}^{2}}\right)\right)^{2} + \tilde{\sigma}_{p}^{2}\mu_{q}^{2} + \sigma_{q}^{2}\tilde{\mu}_{p}^{2} - (\sigma_{q}^{2} + \tilde{\sigma}_{p}^{2})\left(\frac{\sigma_{q}^{2}\tilde{\sigma}_{p}^{2}}{\sigma_{q}^{2} + \tilde{\sigma}_{p}^{2}}\left(\frac{\mu_{q}}{\sigma_{q}^{2}} + \frac{\tilde{\mu}_{p}}{\tilde{\sigma}_{p}^{2}}\right)\right)^{2}}{\sigma_{q}^{2}\tilde{\sigma}_{p}^{2}}.
\end{multline*}
Defining 
\[\tilde{\sigma}_{q}^{2} := \frac{\sigma_{q}^{2}\tilde{\sigma}_{p}^{2}}{\sigma_{q}^{2} + \tilde{\sigma}_{p}^{2}} = \left(\frac{1}{\sigma_{q}^{2}} + \frac{1}{\tilde{\sigma}_{p}^{2}}\right)^{-1} \text{ and } \tilde{\mu}_{q} := \tilde{\sigma}_{q}^{2}\left(\frac{\mu_{q}}{\sigma_{q}^{2}} + \frac{\tilde{\mu}_{p}}{\tilde{\sigma}_{p}^{2}}\right)\]
we get
\begin{equation*}
\begin{split}
q_{\mathbf{\phi}}(\theta)\tilde{p}(\theta)
&= \frac{1}{\theta^{2}2\pi Z_{q}\tilde{Z}_{p}\sqrt{\sigma_{q}^{2}\tilde{\sigma}_{p}^{2}}}\exp\left\{-\frac{1}{2}\frac{(\log\theta - \tilde{\mu}_{q})^{2}}{\tilde{\sigma}_{q}^{2}}\right\}\exp\left\{-\frac{1}{2}\threeTwo\right\}\\
&=\frac{\theta\tilde{Z}_{q}\sqrt{2\pi\tilde{\sigma}_{q}^{2}}}{\theta^{2}2\pi Z_{q}\tilde{Z}_{p}\sqrt{\sigma_{q}^{2}\tilde{\sigma}_{p}^{2}}}\frac{1}{\theta\tilde{Z}_{q}\sqrt{2\pi\tilde{\sigma}_{q}^{2}}}\exp\left\{-\frac{1}{2}\frac{(\log\theta - \tilde{\mu}_{q})^{2}}{\tilde{\sigma}_{q}^{2}}\right\}\exp\left\{-\frac{1}{2}\threeTwo\right\}\\
&=\frac{\tilde{Z}_{q}\sqrt{\tilde{\sigma}_{q}^{2}}}{\theta Z_{q}\tilde{Z}_{p}\sqrt{2\pi\sigma_{q}^{2}\tilde{\sigma}_{p}^{2}}}\exp\left\{-\frac{1}{2}\threeTwo\right\}\text{LogN}_{[a,b]}(\theta|\tilde{\mu}_{q}, \tilde{\sigma}_{q}^{2}).
\end{split}
\end{equation*}

2. Then we divide out $p(\theta)$:
\begin{equation*}
\begin{split}
    q_{\mathbf{\phi}}(\theta)\frac{\tilde{p}(\theta)}{p(\theta)} = \frac{(\log b - \log a)\tilde{Z}_{q}\sqrt{\tilde{\sigma}_{q}^{2}}}{Z_{q}\tilde{Z}_{p}\sqrt{2\pi\sigma_{q}^{2}\tilde{\sigma}_{p}^{2}}}\exp\left\{-\frac{1}{2}\threeTwo\right\}\text{LogN}_{[a,b]}(\theta|\tilde{\mu}_{q}, \tilde{\sigma}_{q}^{2})
\end{split}
\end{equation*}

3. Now we can find $\tilde{q}(\theta)$ and $\Delta F$ using \autoref{eq:dF}. Start with $\tilde{q}(\theta)$:
\begin{equation*}
\begin{split}
    \tilde{q}(\theta) &=  \frac{q_{\mathbf{\phi}}(\theta)\frac{\tilde{p}(\theta)}{p(\theta)}}{\exp \Delta F}\\
    &= \frac{q_{\mathbf{\phi}}(\theta)\frac{\tilde{p}(\theta)}{p(\theta)}}{\int q_{\mathbf{\phi}}(\theta)\frac{\tilde{p}(\theta)}{p(\theta)}d\theta}\\
    &= \frac{\frac{(\log b - \log a)\tilde{Z}_{q}\sqrt{\tilde{\sigma}_{q}^{2}}}{Z_{q}\tilde{Z}_{p}\sqrt{2\pi\sigma_{q}^{2}\tilde{\sigma}_{p}^{2}}}\exp\left\{-\frac{1}{2}\threeTwo\right\}\text{LogN}_{[a,b]}(\theta|\tilde{\mu}_{q}, \tilde{\sigma}_{q}^{2})}{\displaystyle\int \frac{(\log b - \log a)\tilde{Z}_{q}\sqrt{\tilde{\sigma}_{q}^{2}}}{Z_{q}\tilde{Z}_{p}\sqrt{2\pi\sigma_{q}^{2}\tilde{\sigma}_{p}^{2}}}\exp\left\{-\frac{1}{2}\threeTwo\right\}\text{LogN}_{[a,b]}(\theta|\tilde{\mu}_{q}, \tilde{\sigma}_{q}^{2})d\theta}\\
    &= \frac{\frac{(\log b - \log a)\tilde{Z}_{q}\sqrt{\tilde{\sigma}_{q}^{2}}}{Z_{q}\tilde{Z}_{p}\sqrt{2\pi\sigma_{q}^{2}\tilde{\sigma}_{p}^{2}}}\exp\left\{-\frac{1}{2}\threeTwo\right\}\text{LogN}_{[a,b]}(\theta|\tilde{\mu}_{q}, \tilde{\sigma}_{q}^{2})}{ \frac{(\log b - \log a)\tilde{Z}_{q}\sqrt{\tilde{\sigma}_{q}^{2}}}{Z_{q}\tilde{Z}_{p}\sqrt{2\pi\sigma_{q}^{2}\tilde{\sigma}_{p}^{2}}}\exp\left\{-\frac{1}{2}\threeTwo\right\}\int\text{LogN}_{[a,b]}(\theta|\tilde{\mu}_{q}, \tilde{\sigma}_{q}^{2})d\theta}\\
    &= \text{LogN}_{[a,b]}(\theta|\tilde{\mu}_{q}, \tilde{\sigma}_{q}^{2})
\end{split}
\end{equation*}

4. Finally we get $\Delta F$:
\begin{equation*}
\begin{split}
    \Delta F &= \log \frac{q_{\mathbf{\phi}}(\theta)\frac{\tilde{p}(\theta)}{p(\theta)}}{\tilde{q}(\theta)} \\
    &= \log \frac{(\log b - \log a)\tilde{Z}_{q}\sqrt{\tilde{\sigma}_{q}^{2}}}{Z_{q}\tilde{Z}_{p}\sqrt{2\pi\sigma_{q}^{2}\tilde{\sigma}_{p}^{2}}}\exp\left\{-\frac{1}{2}\threeTwo\right\}\\
    &= \log \frac{\tilde{Z}_{q}(\log b - \log a)}{Z_{q}\tilde{Z}_{p}} + \frac{1}{2}\log \frac{\tilde{\sigma}_{q}^{2}}{2\pi\sigma_{q}^{2}\tilde{\sigma}_{p}^{2}} - \frac{1}{2}\threeTwo
\end{split}
\end{equation*}

\subsection{\methodu{}}
\label{sec:bmrsu_derivation}
The PDF of the reduced truncated log-uniform distribution is given as follows:
\begin{equation}
    \tilde{p}(\theta) = \text{LogU}_{[a', b']}(\theta) = \begin{cases}
    \left({\theta\log \frac{b'}{a'}}\right)^{-1}, & a \le a' < b' \le b \\
    0, & \text{otherwise}
  \end{cases}
\end{equation}
Using this, we can directly solve the integral under the expectation given in \autoref{eq:dF} for $\Delta F$
\begin{equation*}
    \exp \Delta F = \mathbb{E}_{\tilde{p}}\left[\frac{q_{\mathbf{\phi}}(\theta)}{p(\theta)}\right] = \int_{a}^{b}\text{LogU}_{[a',b']}(\theta)\frac{q_\mathbf{\phi}(\theta)}{\text{LogU}_{[a,b]}(\theta)} d\theta = \int_{a}^{b}\frac{\theta\log \frac{b}{a}}{\theta\log \frac{b'}{a'}} q_\mathbf{\phi}(\theta)d\theta
\end{equation*}
\begin{equation*}
 = \int_{a}^{a'}0d\theta + \int_{a'}^{b'}\frac{\log\frac{b}{a}}{\log \frac{b'}{a'}}q_\mathbf{\phi}(\theta)d\theta + \int_{b'}^{b}0d\theta 
= \frac{\log\frac{b}{a}}{\log \frac{b'}{a'}}q_{\mathbf{\phi}}(a' \le \theta_{i} \le b')
\end{equation*}
Plugging this in to \autoref{eq:dF_derivation} where $\exp \Delta F \ge 1$:
\begin{equation}
    1 \le \frac{\log\frac{b}{a}}{\log \frac{b'}{a'}}q_{\mathbf{\phi}}(a' \le \theta_{i} \le b')
\Rightarrow
    \frac{\log\frac{b'}{a'}}{\log \frac{b}{a}} \le q_{\mathbf{\phi}}(a' \le \theta \le b')
\end{equation}
\section{Dataset details}
\label{sec:dataset_details}

\paragraph{MNIST} MNIST~\cite{lecun1998mnist} is a classic image classification dataset consisting of 70,000 28x28 black and white images of handwritten digits (10 classes). We use the original 10,000 image test set for testing and split the 60,000 image train set into 80\% training and 20\% validation images.
\paragraph{Fashion-MNIST} Fashion-MNIST~\cite{DBLP:journals/corr/abs-1708-07747} is a modernized version of MNIST using images of different articles of clothing as opposed to handwritten digits. The dataset statistics are the same as MNIST: 28x28 greyscale images, 60,000 training images, 10,000 test images, 10 classes. Similar to MNIST, we split the training set into 80\% training and 20\% validation images.
\paragraph{CIFAR10} CIFAR10~\cite{krizhevsky2009learning} is an image classification dataset of 32x32 color images with 10 classes. There are 50,000 training images and 10,000 test images. Again, we split the training set to 80\% training and 20\% validation.
\paragraph{TinyImagenet} TinyImagenet is a reduced version of ImageNet consisting of 110,000 64x64 color images in 200 classes. We use the 10,000 image validation split for testing, and split the 100,000 image train set into 80\% training and 20\% validation images.

\section{Model details}
\label{sec:model_details}

For each model and dataset we use the Adam optimizer with no weight decay. We train for 50 epochs for each experiment with an MLP and Lenet5, and for 100 epochs for each experiment with Resnet50 and ViT. Further details about each model are given as follows:

\subsection{MLP} 
We use a multilayer perceptron (MLP) for several experiments, with different network sizes based on a hyperparameter sweep for each dataset. Multiplicative noise for pruning is applied to every neuron in the network. We sweep through the following hyperparmeters:

\begin{itemize}[leftmargin=*]
\setlength{\itemsep}{0pt}
    \item Number of layers = \{1,3,5,7,9\}
    \item Hidden dimension = \{10, 30, 50, 100, 150\}
    \item Batch size = \{16, 32, 64, 128\}
    \item Learning rate [0.0001, 0.1].
\end{itemize}

The final network settings for each dataset are given as follows:

\begin{description}
\item[MNIST:] Number of layers: 7; Hidden dimension: 100; Batch size: 128; Learning rate: 8.5$\cdot10^{-4}$.
\item[Fashion-MNIST] Number of layers: 1; Hidden dimension: 150; Batch size: 128; Learning rate: 1.5$\cdot10^{-3}$.
\item[CIFAR10] Number of layers: 5; Hidden dimension: 150; Batch size: 32; Learning rate: 6.8$\cdot10^{-4}$.
\end{description}

\subsection{Lenet5}
Lenet5~\cite{DBLP:journals/pieee/LeCunBBH98} is an early CNN architecture consisting of 2 convolutional layers with 6 and 16 filters per channel, respectively, each followed by a ReLU activation and max pooling layer, followed by 3 linear layers. Multiplicative noise is applied to each convolutional filter map, as well as each neuron the the linear layers. We use the same architecture for each experiments and tune hyperparameters based on the dataset. We sweep through the following hyperparameters:

\begin{itemize}[leftmargin=*]
\setlength{\itemsep}{0pt}
    \item Batch size = \{16, 32, 64, 128\}
    \item Learning rate [0.0001, 0.1].
\end{itemize}

The final settings for each dataset are given as follows:

\begin{description}
\item[MNIST:] Batch size 128; Learning rate 1.4$\cdot10^{-3}$.
\item[Fashion-MNIST] Batch size 32; Learning rate 1.4$\cdot10^{-3}$.
\item[CIFAR10] Batch size 64; Learning rate 1$\cdot10^{-3}$.
\end{description}

\subsection{Resnet50}
Resnet50~\cite{DBLP:conf/cvpr/HeZRS16} is a deep 50-layer CNN which uses residual connections to stabilize optimization and improve accuracy. We start with a model pretrained on ImageNet-1k,\footnote{\url{https://pytorch.org/vision/stable/models.html}} then fine-tuned on the downstream dataset with pruning layers added to each output layer after batch normalization. We use a learning rate of 6.8e-4, a batch size of 32, and train for 100 epochs for both CIFAR10 and TinyImagenet.

\subsection{Vision Transformer (ViT)}
Vision Transformer (ViT)~\cite{wu2020visual} is a transformer model tailored for image data based on tokenizing an image as 16x16 image patches. We use a ViT which is pretrained on ImageNet-21k (14M images and 21,843 classes) as well as ImageNet-1k.\footnote{\url{https://huggingface.co/google/vit-base-patch16-224}} We add multiplicative noise to the output layer of each transformer block for pruning. We use a learning rate of 6.8e-4, a batch size of 32, and train for 100 epochs for both CIFAR10 and TinyImagenet.

\section{Compute resources}
\label{sec:compute_resources}
All experiments were run on a shared cluster. Requested jobs consisted of 16GB of RAM and 4
Intel Xeon Silver 4110 CPUs. We used a single
NVIDIA Titan X GPU with 24GB of RAM for all experiments, though utilization was generally much lower due the the average size of each network. Runtimes for each experiment ranged from approx. 7 minutes for Lenet5 on MNIST with no pruning layers to approx. 44 hours for ViT on TinyImagenet with multiplicative noise trained for 100 epochs. The training of models in this work over the course of the entire project (prototyping, experimentation, etc.) is estimated to have used 3773.785 kWh of electricity contributing to 599.892 kg of CO2eq (as measured by carbontracker~\cite{anthony2020carbontracker}; this is equivalent to 5580.395 km travelled by car).

We additionally benchmark the training and inference runtimes of each model \textit{before} pruning in \autoref{tab:small_timing_results} and \autoref{tab:large_timing_results} (i.e., the inference runtimes are without removing structures).

\begin{table}%[htp]
    \def\arraystretch{1.2}
    \centering
    %\fontsize{10}{10}\selectfont
    \scriptsize
    %\rowcolors{2}{gray!10}{white}
    \caption{Training and inference runtimes in milliseconds. Each runtime is averaged across 1000 forward passes of a batch of 32 images. Inference runtimes are \textit{before} pruning (i.e., with the full network).} %Results are reported as the average macro F1 over 5 random seeds.}
    \begin{tabular}{l c c | c c | c c}
    \toprule %\thickhline
    & \multicolumn{2}{c}{ MNIST } & \multicolumn{2}{c}{ Fash-MNIST } & \multicolumn{2}{c}{ CIFAR10 } \\
    \midrule
    Pruning Method & Train & Inf. & Train & Inf. & Train & Inf.\\

\midrule
\multicolumn{7}{c}{MLP}\\
\midrule
None&$2.41$& $0.73$& $2.43$& $0.73$& $4.61$& $5.30$\\
$E[\theta]$&$19.76$& $1.17$& $20.00$& $1.14$& $21.21$& $5.39$\\
SNR&$20.00$& $1.14$& $19.99$& $1.15$& $22.00$& $4.94$\\
\methodn{}&$20.02$& $1.14$& $20.14$& $1.15$& $21.76$& $5.57$\\
\methodu{}&$19.99$& $1.16$& $20.11$& $1.16$& $22.14$& $5.15$\\
\midrule
\multicolumn{7}{c}{Lenet5}\\
\midrule
None&$2.31$& $0.69$& $2.24$& $0.69$& $4.73$& $5.15$\\
$E[\theta]$&$11.45$& $0.94$& $11.34$& $0.94$& $12.11$& $4.90$\\
SNR&$11.18$& $0.93$& $11.42$& $0.94$& $15.05$& $4.95$\\
\methodn{}&$11.39$& $0.94$& $11.47$& $0.94$& $13.34$& $5.07$\\
\methodu{}&$11.44$& $0.94$& $11.08$& $0.93$& $12.86$& $4.43$\\

    \bottomrule % \thickhline

    \end{tabular}
    
    \label{tab:small_timing_results}
\end{table}

\begin{table}%[htp]
    \def\arraystretch{1.2}
    \centering
    %\fontsize{10}{10}\selectfont
    \scriptsize
    %\rowcolors{2}{gray!10}{white}
    \caption{Training and inference runtimes in milliseconds. Each runtime is averaged across 1000 forward passes of a batch of 32 images. Inference runtimes are \textit{before} pruning (i.e., with the full network).} %Results are reported as the average macro F1 over 5 random seeds.}
    \begin{tabular}{l c c | c c}
    \toprule %\thickhline
    & \multicolumn{2}{c}{ CIFAR10 } & \multicolumn{2}{c}{ TinyImagenet } \\
    \midrule
    Pruning Method & Train & Inf. & Train & Inf.\\

\midrule
\multicolumn{5}{c}{Res50-Pretrained}\\
\midrule
None&$47.78$& $34.30$& $74.89$& $59.11$\\
$E[\theta]$&$329.11$& $35.24$& $346.80$& $64.02$\\
SNR&$328.29$& $33.97$& $346.56$& $64.00$\\
\methodn{}&$329.84$& $32.99$& $345.63$& $64.01$\\
\methodu{}-4&$329.81$& $34.85$& $346.78$& $64.06$\\
\midrule
\multicolumn{5}{c}{Vision Transformer}\\
\midrule
None&$254.50$& $87.15$& $252.91$& $86.70$\\
$E[\theta]$&$343.51$& $88.80$& $351.19$& $87.49$\\
SNR&$343.50$& $88.92$& $350.93$& $87.55$\\
\methodn{}&$343.55$& $88.74$& $350.99$& $87.55$\\
\methodu{}-4&$343.60$& $88.56$& $351.39$& $87.47$\\

    \bottomrule % \thickhline

    \end{tabular}
    
    \label{tab:large_timing_results}
\end{table}

\section{Additional plots}
\label{sec:additional_plots}
\begin{figure}[t!]
    \centering
    \begin{subfigure}[t]{0.498\textwidth}
        \centering
    \includegraphics[width=0.48\textwidth]{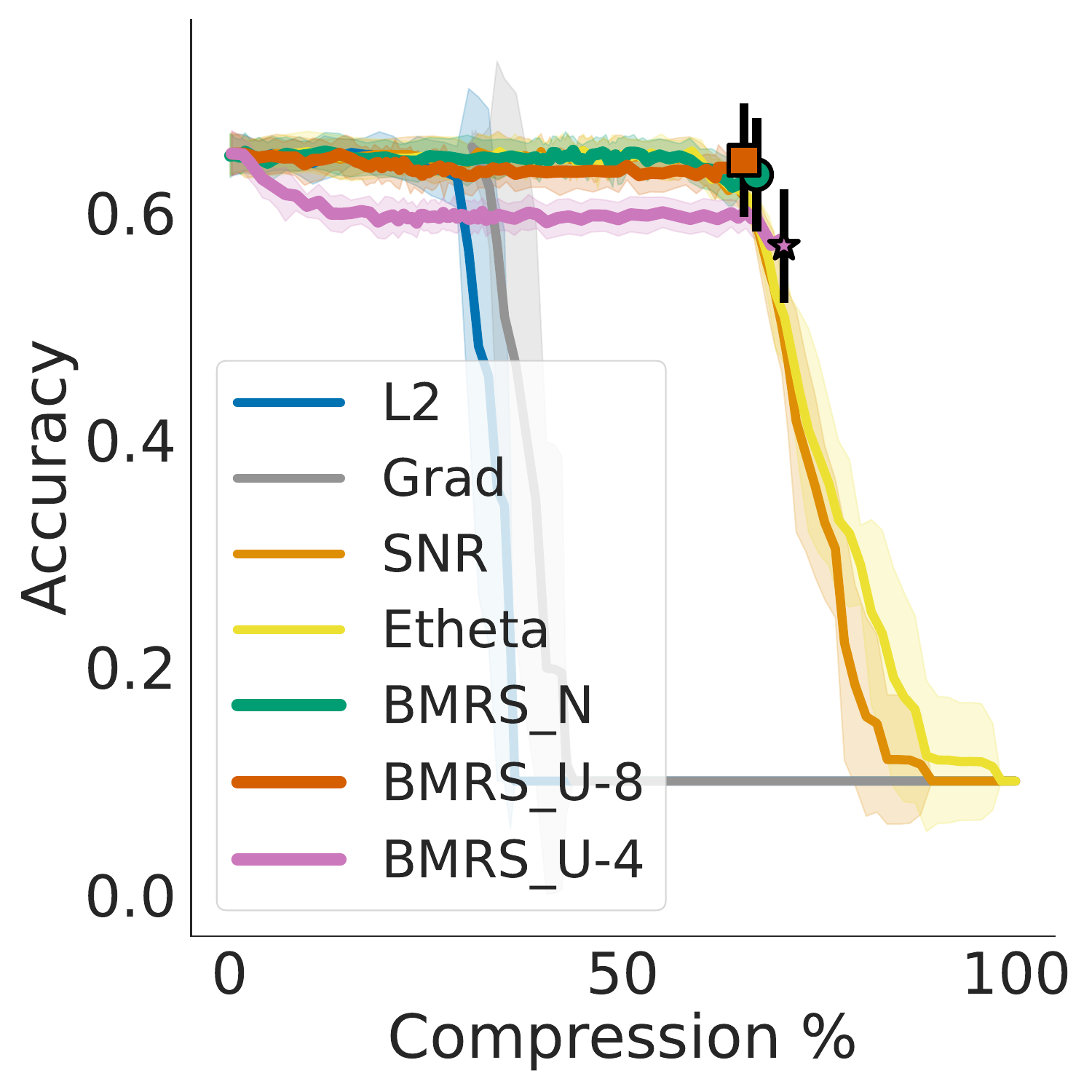}
    \includegraphics[width=0.48\textwidth]{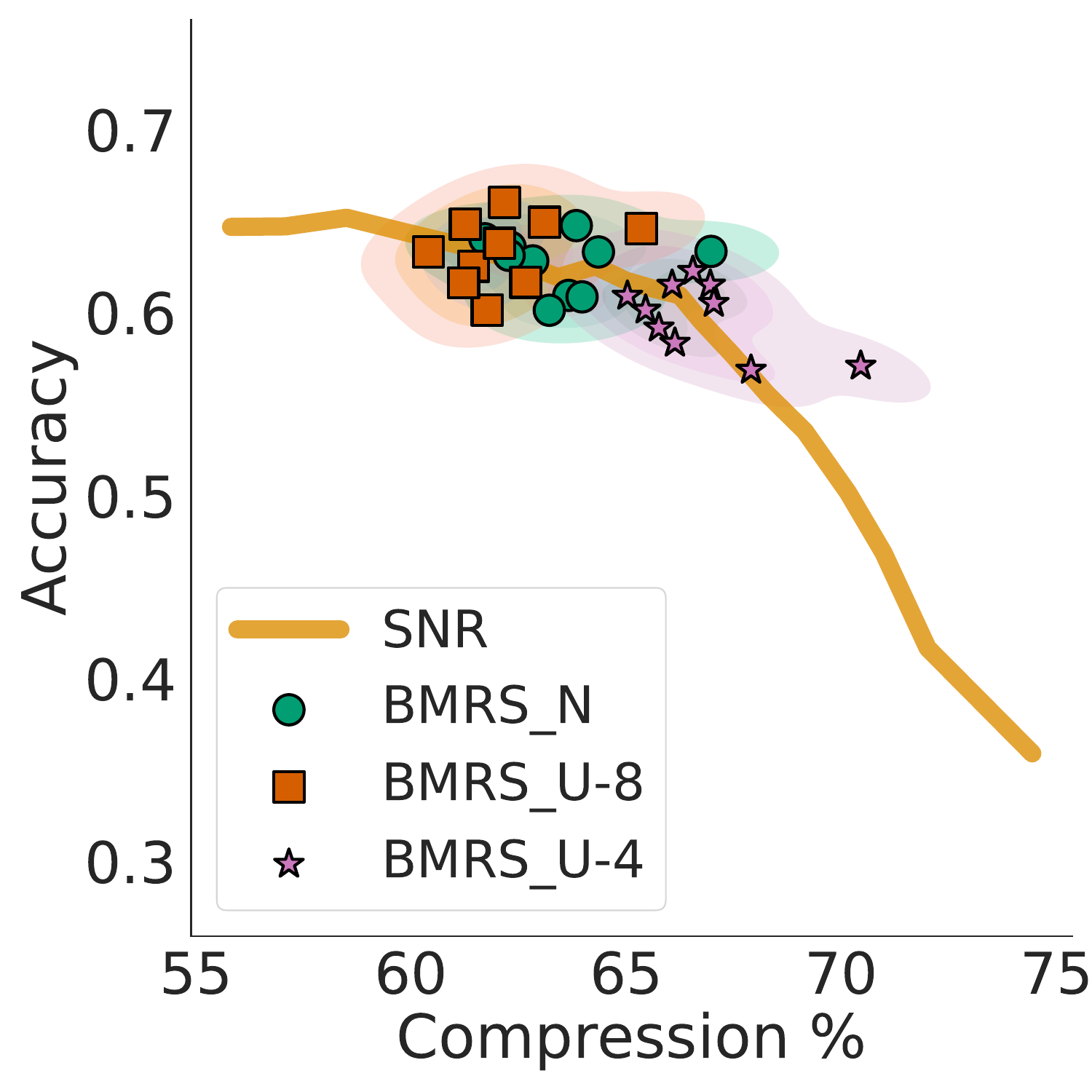}
        \caption{CIFAR10 Lenet5}
    \end{subfigure}%
    ~
    \begin{subfigure}[t]{0.498\textwidth}
        \centering
        \includegraphics[width=0.48\textwidth]{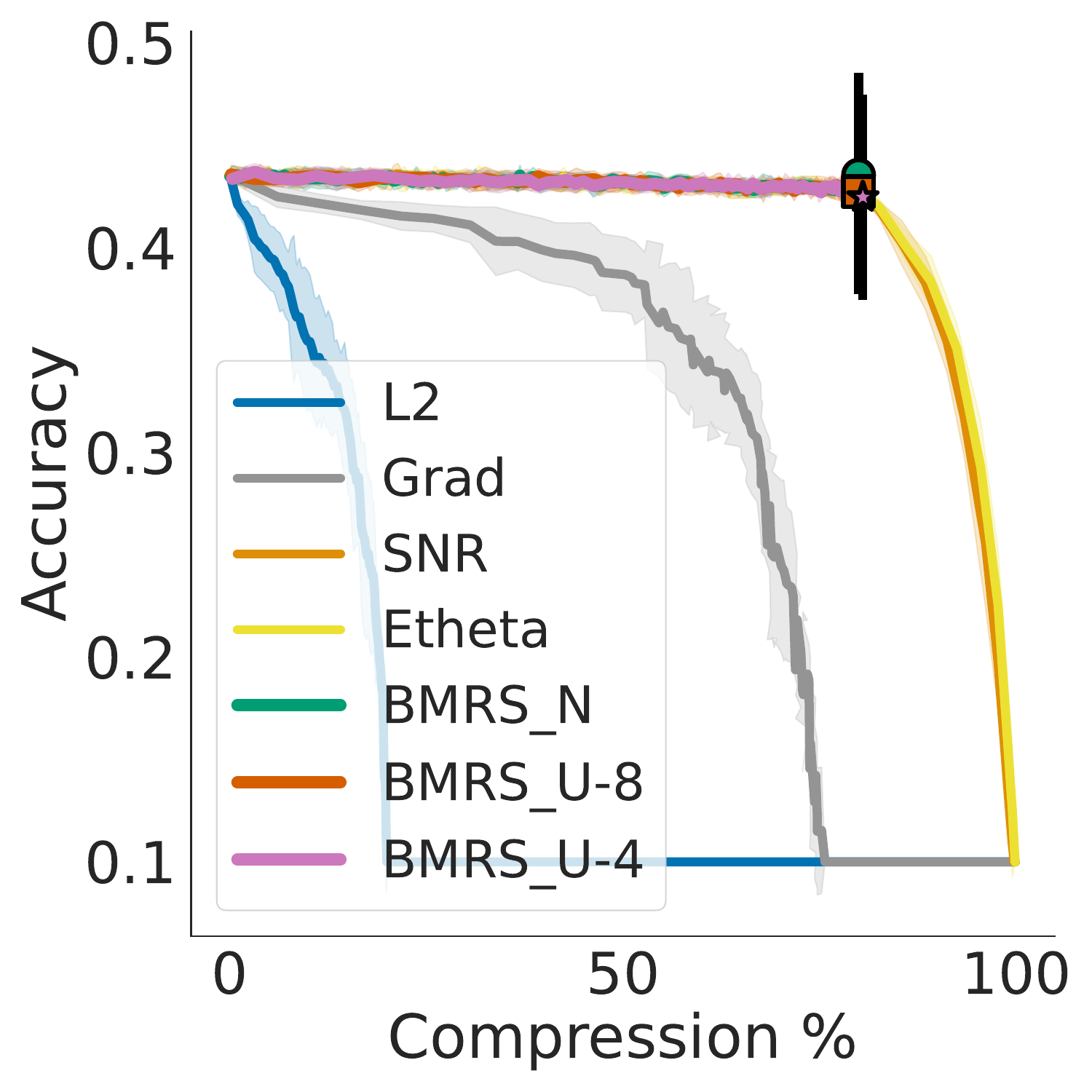}
        \includegraphics[width=0.48\textwidth]{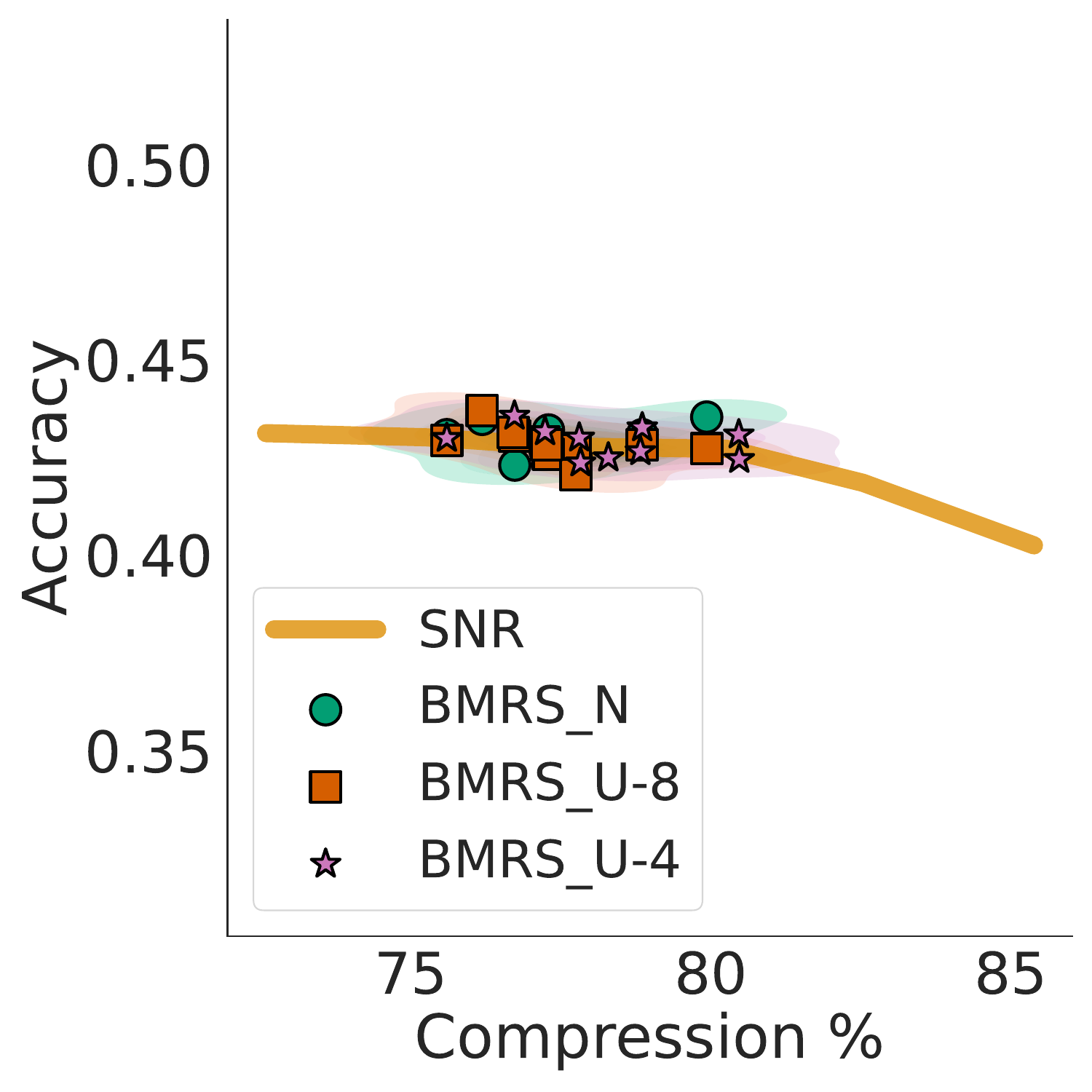}
        \caption{CIFAR10 MLP}
    \end{subfigure}% 
    \\
    \begin{subfigure}[t]{0.498\textwidth}
        \centering
    \includegraphics[width=0.48\textwidth]{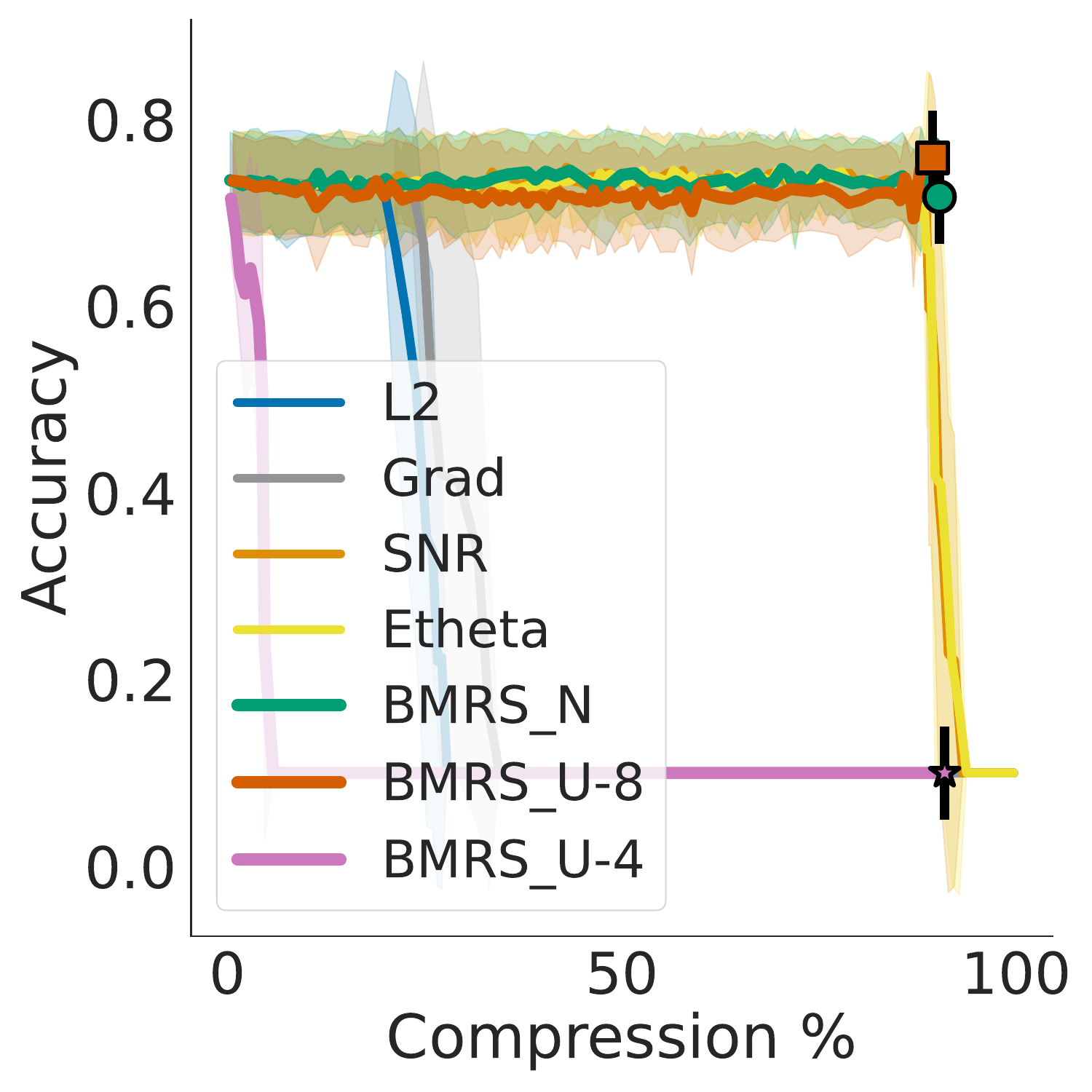}
    \includegraphics[width=0.48\textwidth]{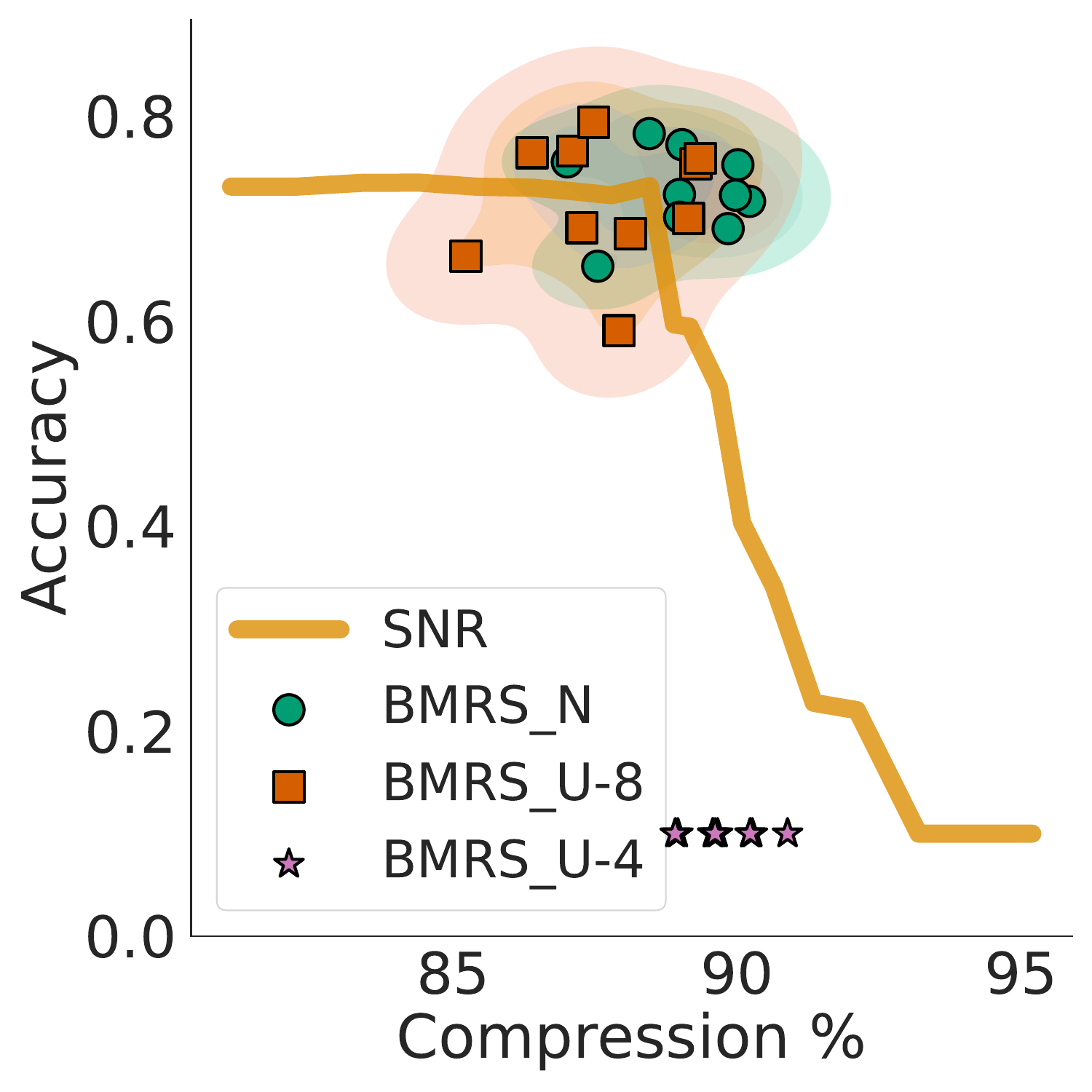}
        \caption{Fashion-MNIST Lenet5}
    \end{subfigure}%
    ~
    \begin{subfigure}[t]{0.498\textwidth}
        \centering
        \includegraphics[width=0.48\textwidth]{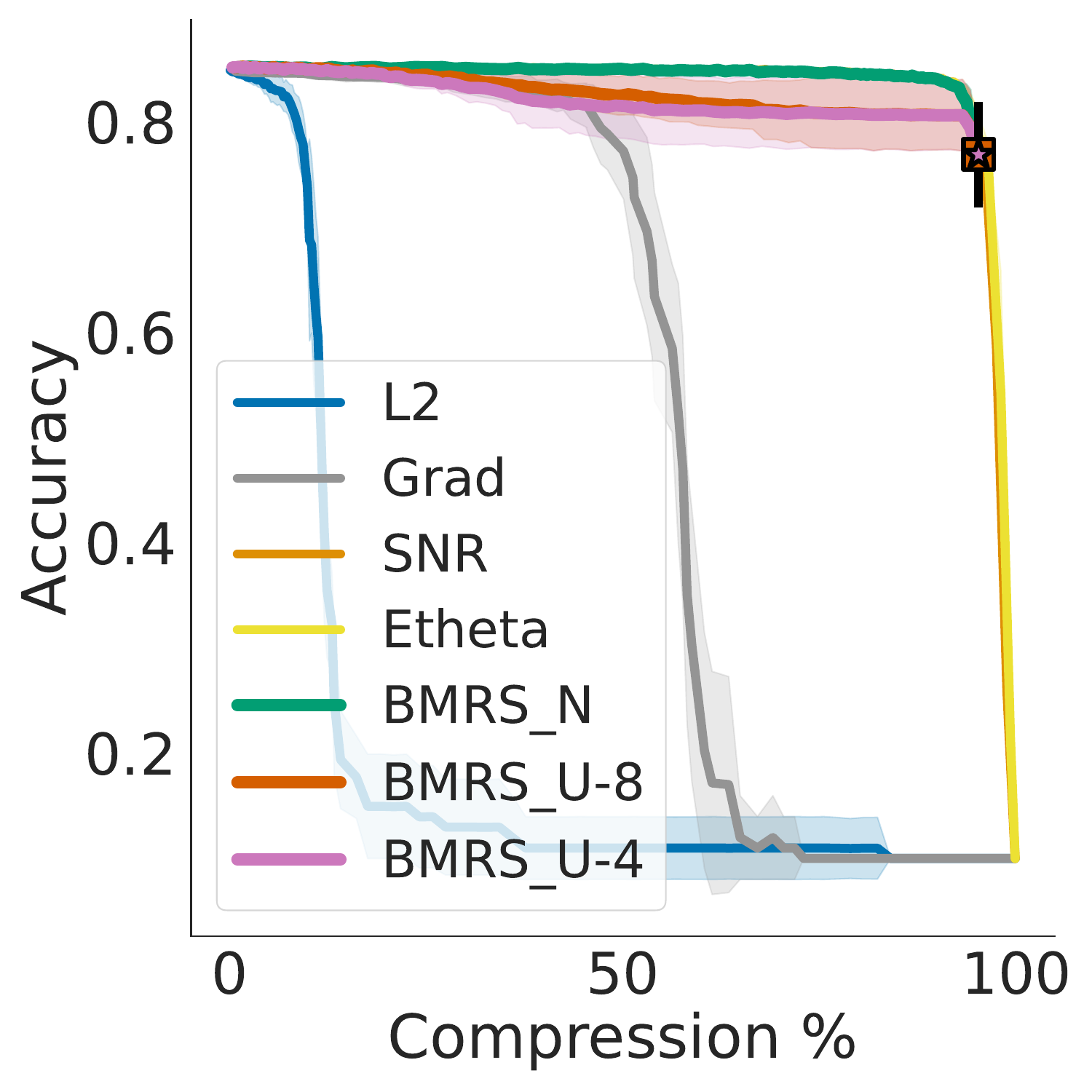}
        \includegraphics[width=0.48\textwidth]{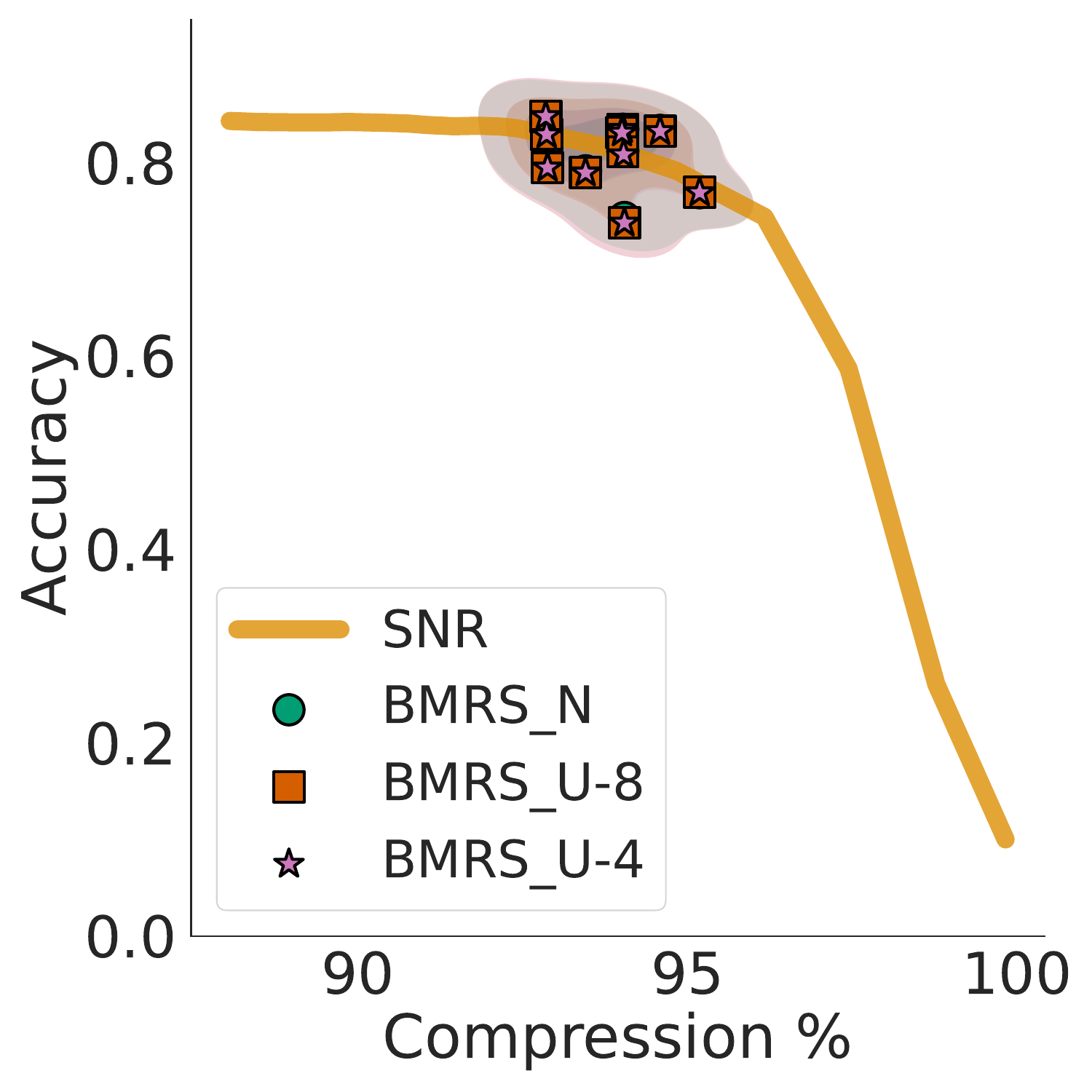}
        \caption{Fashion-MNIST MLP}
    \end{subfigure}%
    \\
    \begin{subfigure}[t]{0.498\textwidth}
        \centering
    \includegraphics[width=0.48\textwidth]{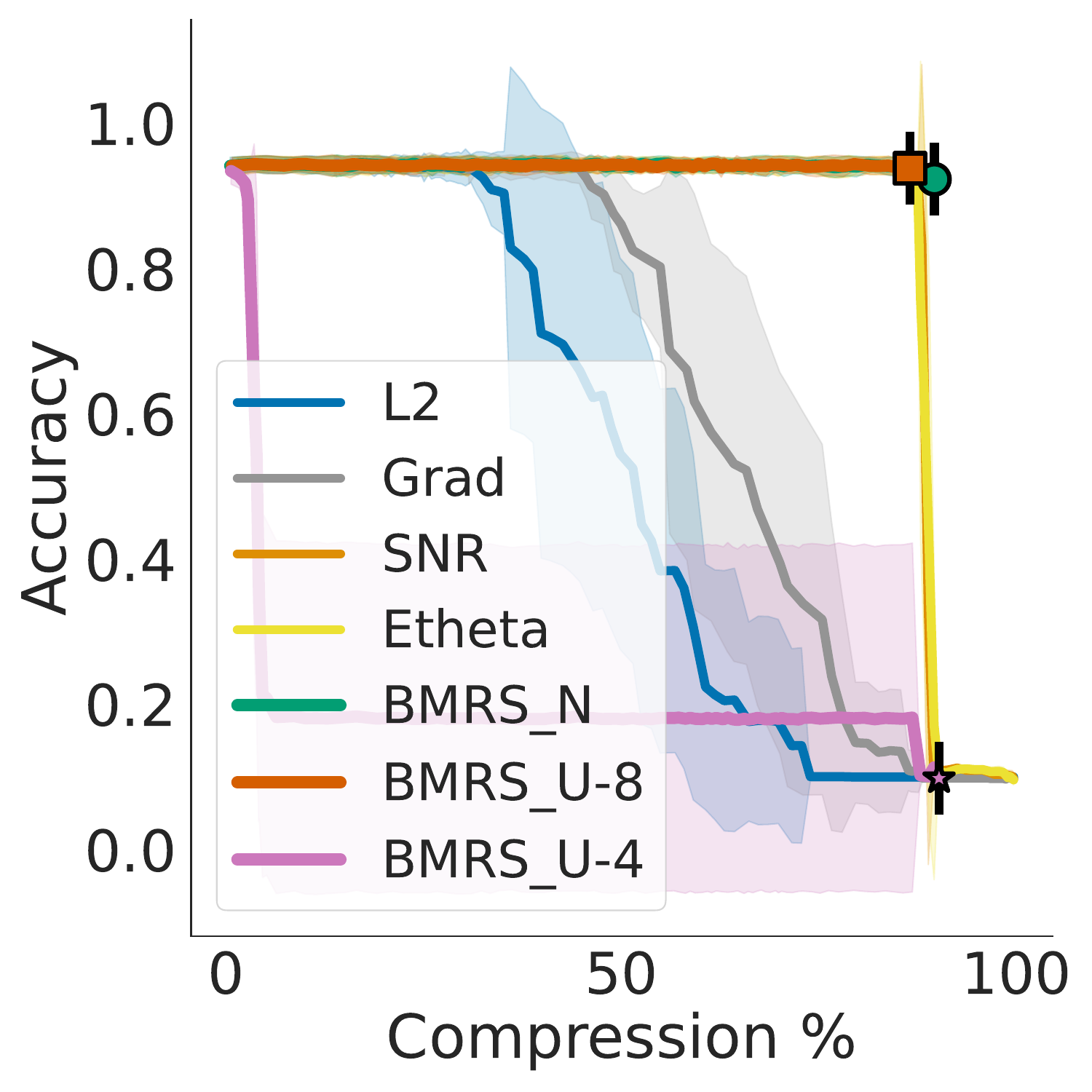}
    \includegraphics[width=0.48\textwidth]{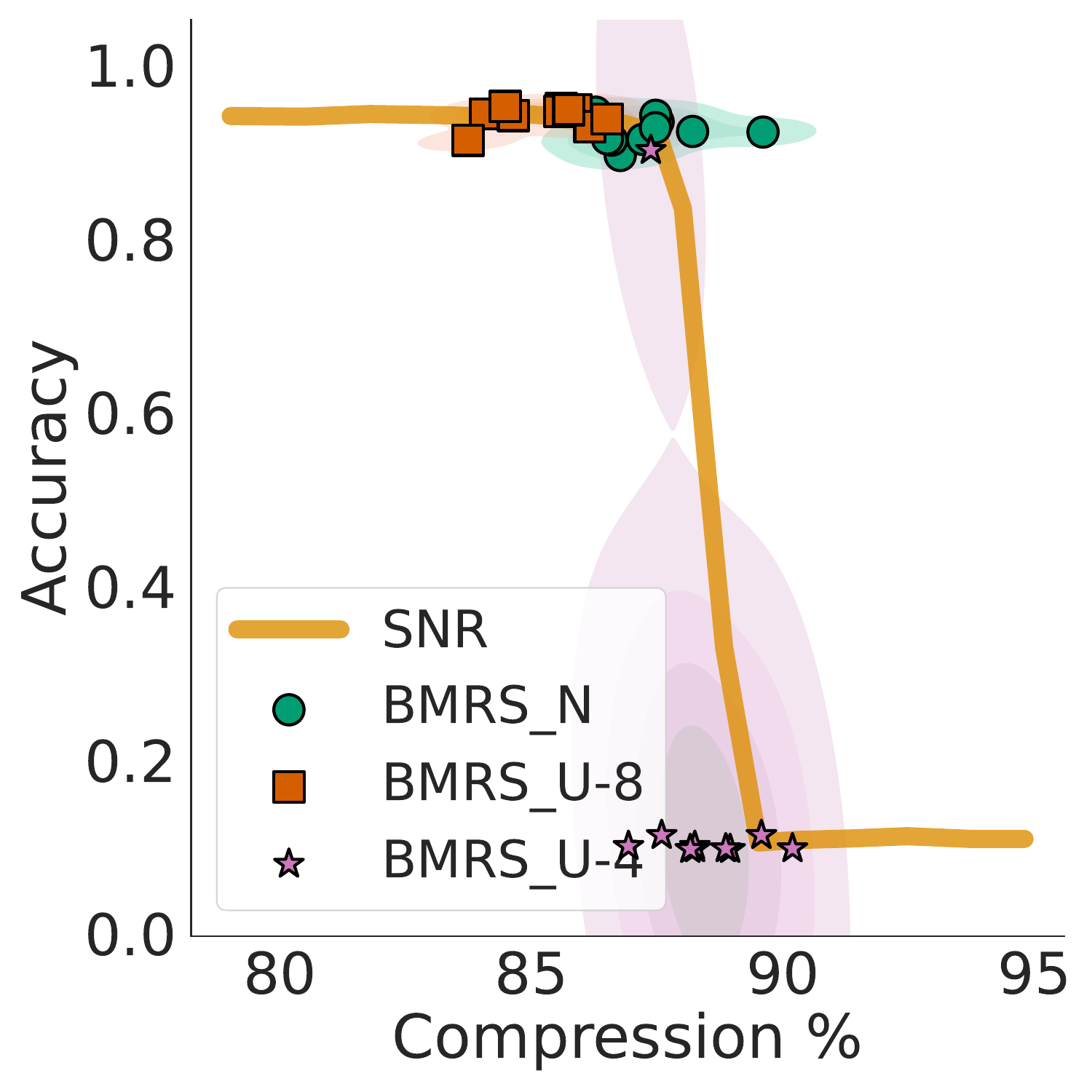}
        \caption{MNIST Lenet5}
    \end{subfigure}%
    ~
    \begin{subfigure}[t]{0.498\textwidth}
        \centering
        \includegraphics[width=0.48\textwidth]{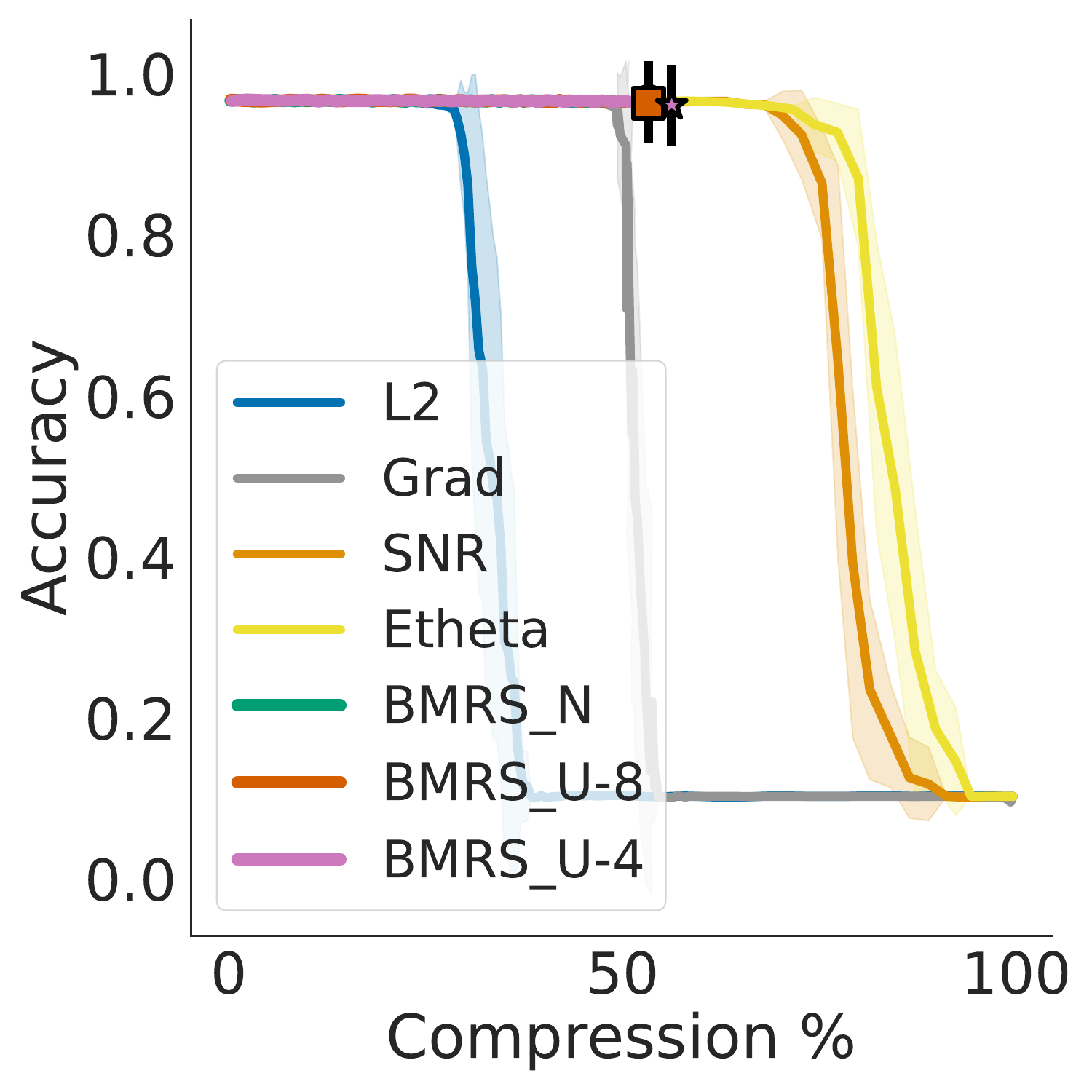}
        \includegraphics[width=0.48\textwidth]{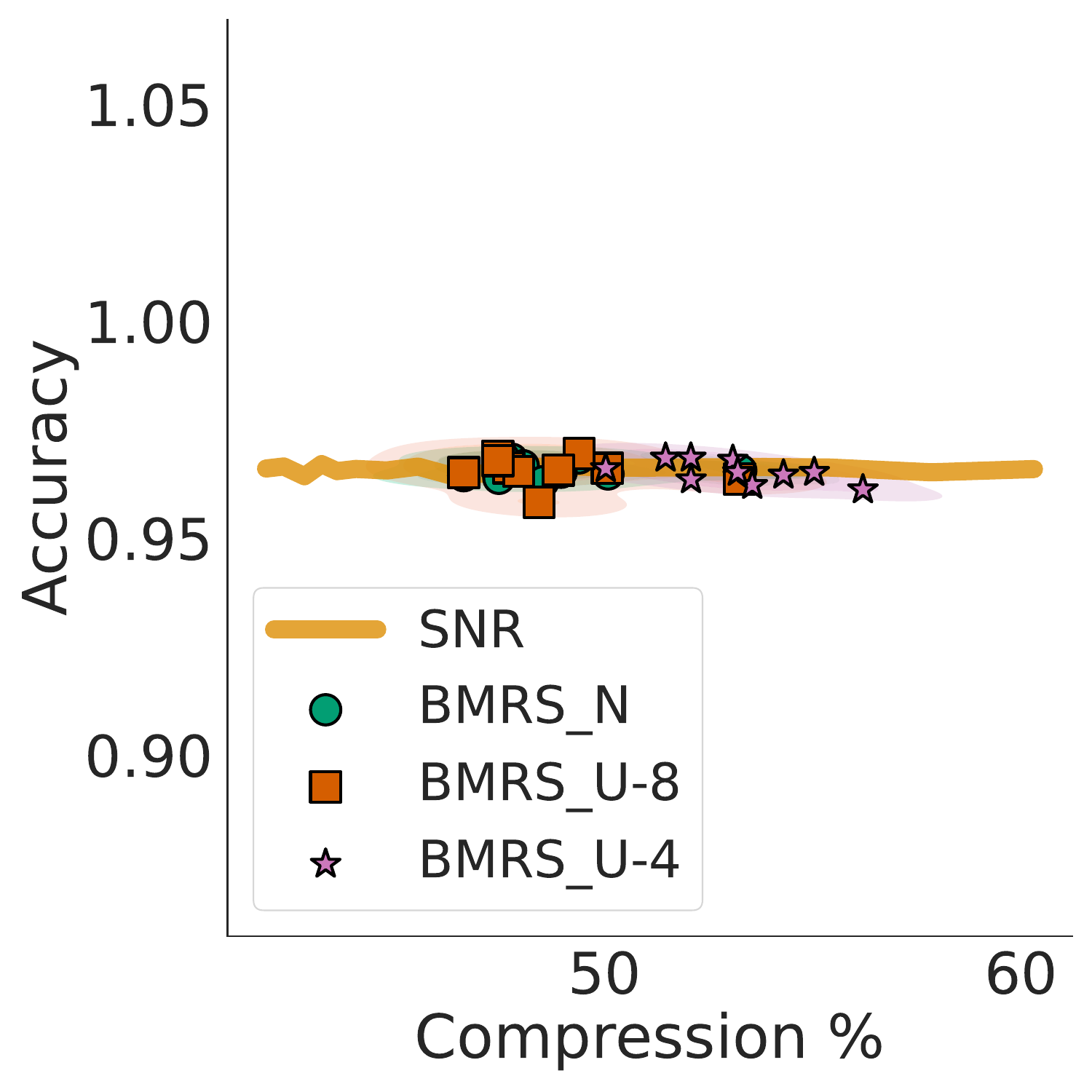}
        \caption{MNIST MLP}
    \end{subfigure}%
    \caption{Additional accuracy vs.~compression results for post-training pruning including gradient-based pruning.}
    \label{fig:post_training_graphs_supplement}

\end{figure}

An additional experiment for post-training pruning including gradient-based thersholding for pruning is given in \autoref{fig:post_training_graphs_supplement}. For gradient based pruning, we use two thresholds: one for the L2-Norm (magnitude) of structures, and one for the L2-Norm of the gradients of weights in a given structure.

\begin{figure*}[h]
    \centering
    \begin{subfigure}[t]{0.35\textwidth}
        \centering
    \includegraphics[width=\textwidth]{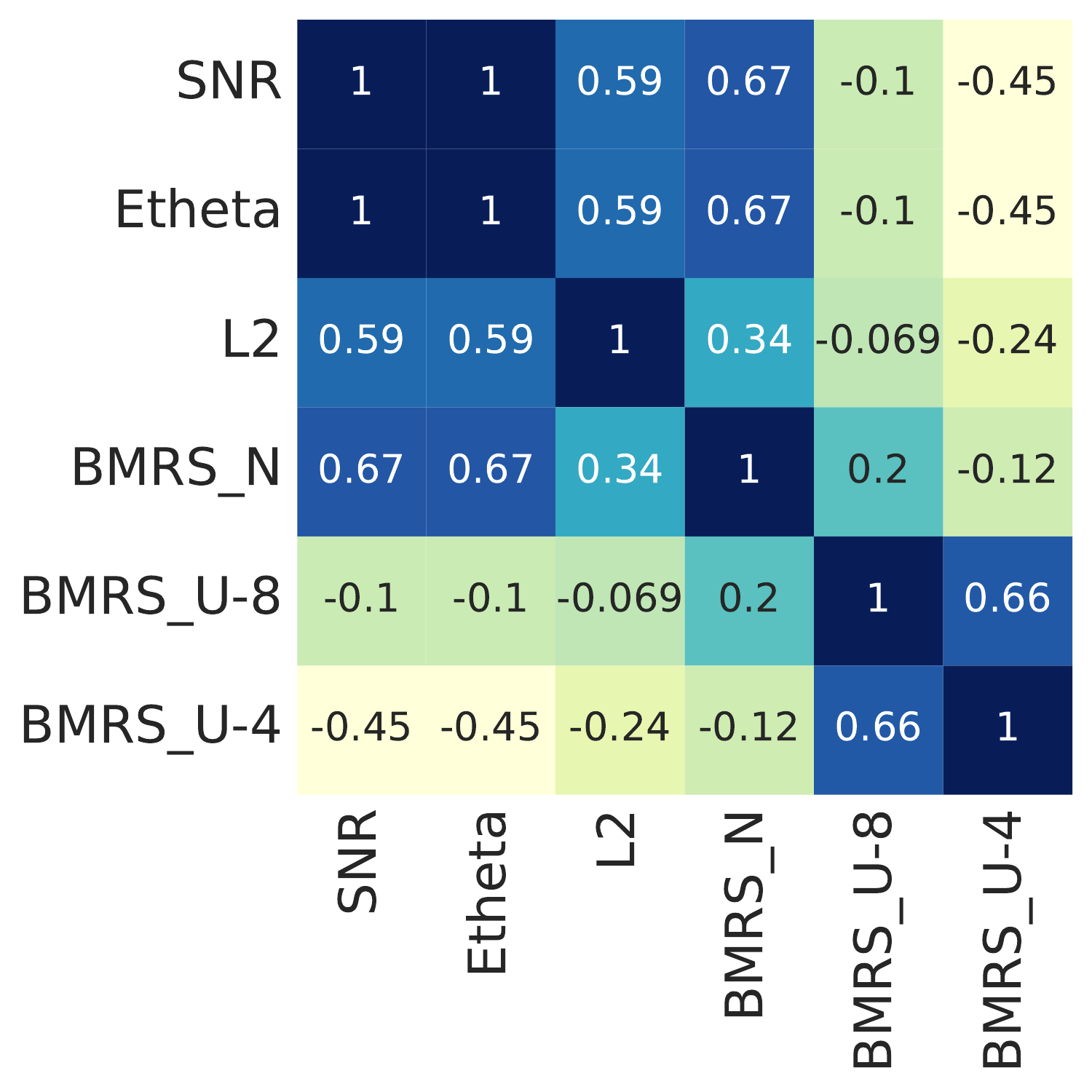}
        \caption{Fashion-MNIST Lenet5}
    \end{subfigure}%
    ~
    \begin{subfigure}[t]{0.35\textwidth}
        \centering
        \includegraphics[width=\textwidth]{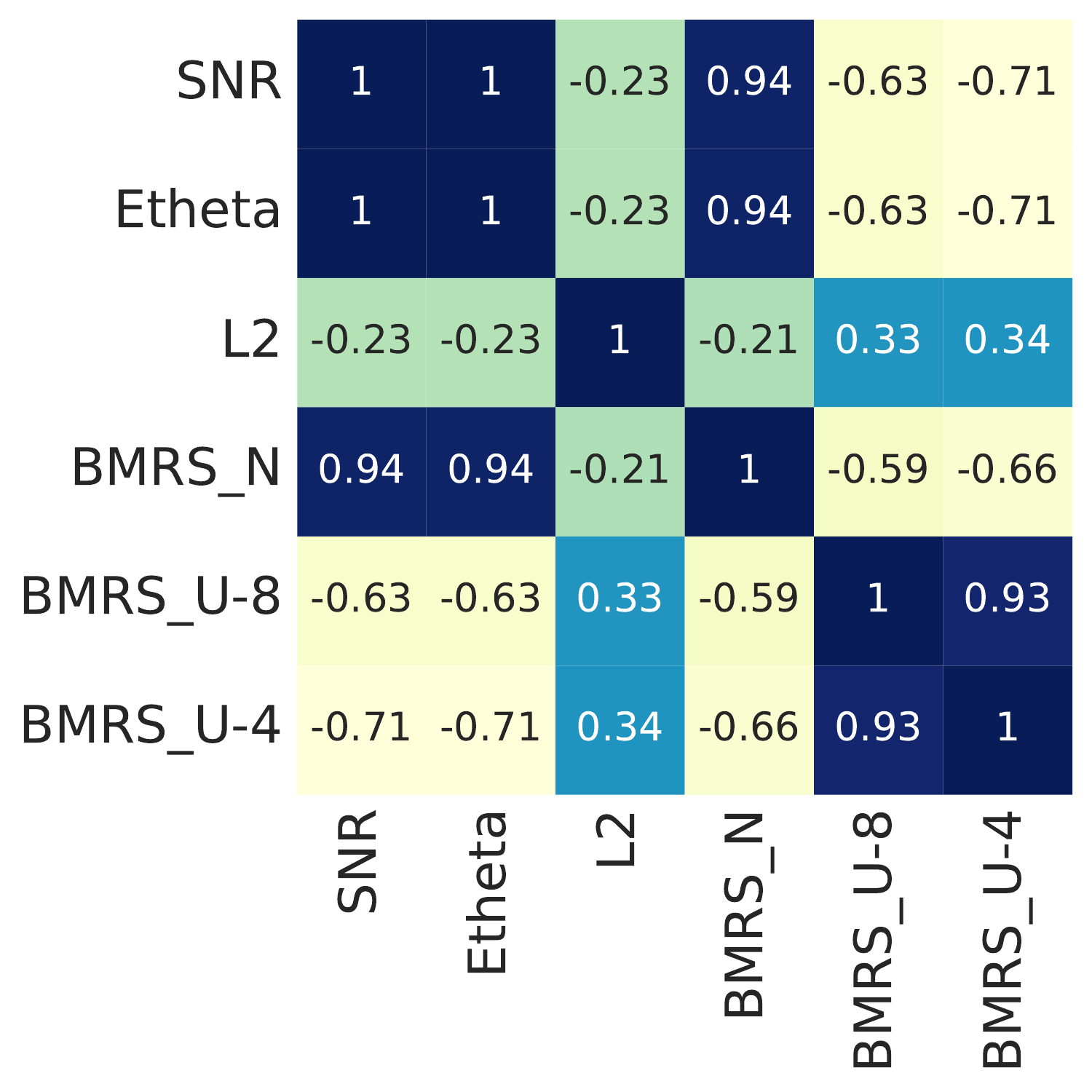}
        \caption{Fashion-MNIST MLP}
    \end{subfigure}%
    \caption{Average correlation between the ranks of neurons for pruning when using different methods on Fashion-MNIST.}
    \label{fig:correlations_fashion_mnist}
\end{figure*}

\begin{figure*}[h]
    \centering
    \begin{subfigure}[t]{0.35\textwidth}
        \centering
    \includegraphics[width=\textwidth]{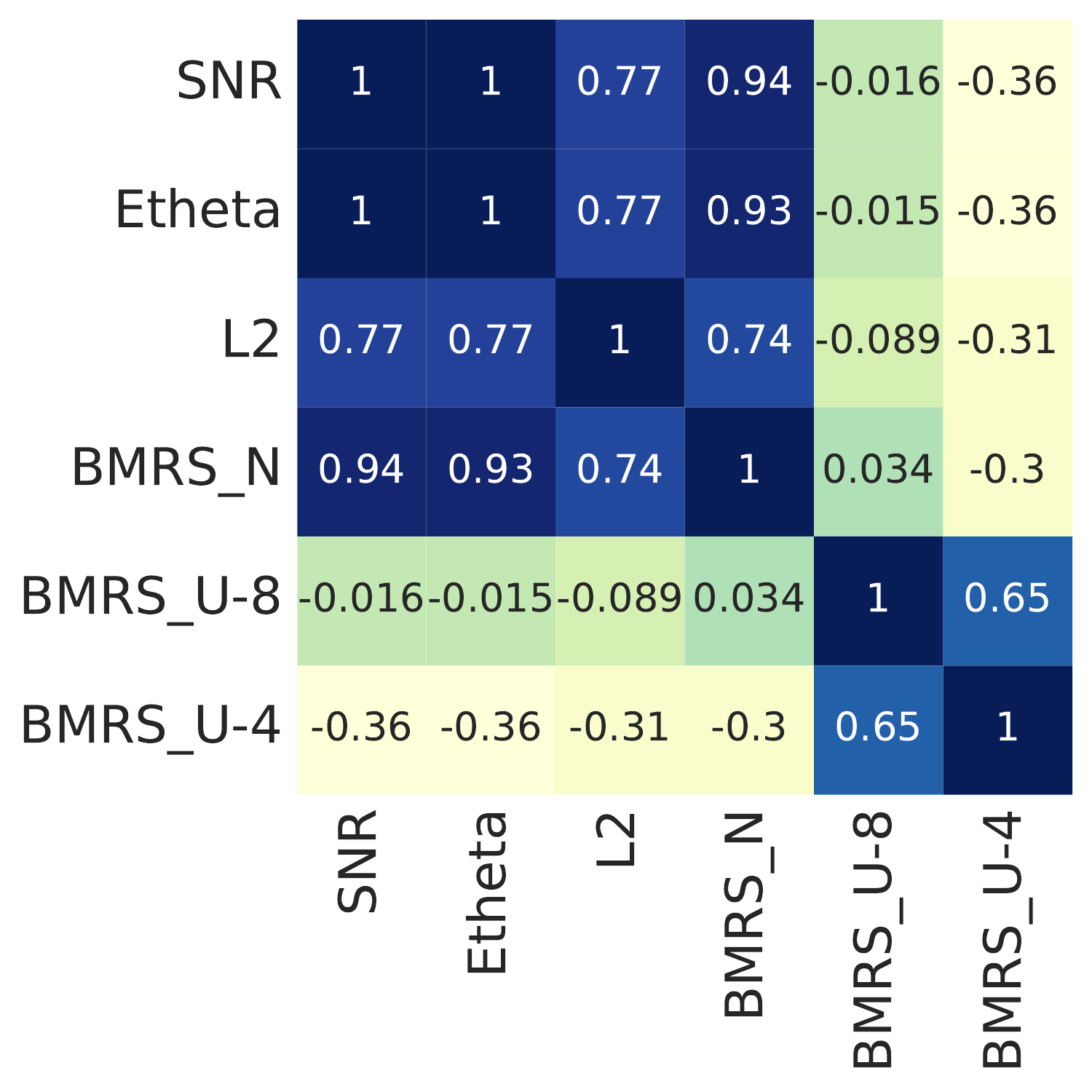}
        \caption{MNIST Lenet5}
    \end{subfigure}%
    ~
    \begin{subfigure}[t]{0.35\textwidth}
        \centering
        \includegraphics[width=\textwidth]{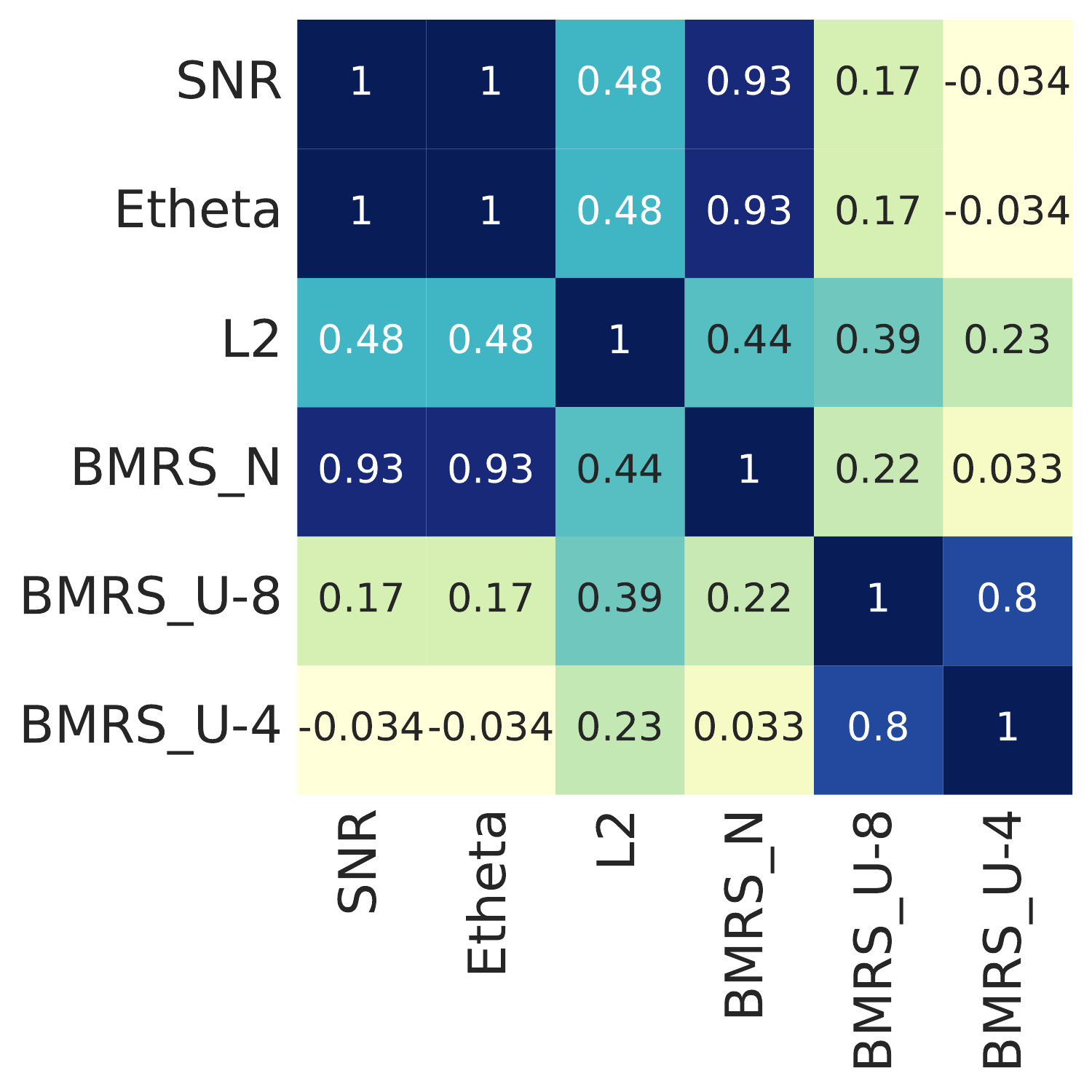}
        \caption{MNIST MLP}
    \end{subfigure}%
    \caption{Average correlation between the ranks of neurons for pruning when using different methods on MNIST.}
    \label{fig:correlations_mnist}
\end{figure*}

The neuron rank correlations for Fashion-MNIST are given in \autoref{fig:correlations_fashion_mnist}. and for MNIST in \autoref{fig:correlations_mnist}.

% \begin{figure*}[h]
%     \centering
%     \begin{subfigure}[t]{0.49\textwidth}
%         \centering
%     \includegraphics[width=0.48\textwidth]{neurips_rebuttal/fashion_mnist_lenet5_cva_combined.pdf}
%     \includegraphics[width=0.48\textwidth]{neurips_rebuttal/fashion_mnist_lenet5_cva_zoomed_combined.pdf}
%         \caption{Fashion-MNIST Lenet5}
%     \end{subfigure}%
%     ~
%     \begin{subfigure}[t]{0.49\textwidth}
%         \centering
%         \includegraphics[width=0.48\textwidth]{neurips_rebuttal/fashion_mnist_mlp_cva_combined.pdf}
%         \includegraphics[width=0.48\textwidth]{neurips_rebuttal/fashion_mnist_mlp_cva_zoomed_combined.pdf}
%         \caption{Fashion-MNIST MLP}
%     \end{subfigure}%
%     \caption{Accuracy vs. compression for pruning when using different methods on Fashion-MNIST.}
%     \label{fig:post_training_graphs_fashion_mnist}
    
% \end{figure*}

% \begin{figure*}[h]
% \label{fig:post_training_graphs_supplement}
%     \centering
%     \begin{subfigure}[t]{0.49\textwidth}
%         \centering
%     \includegraphics[width=0.48\textwidth]{neurips_rebuttal/mnist_lenet5_cva_combined.pdf}
%     \includegraphics[width=0.48\textwidth]{neurips_rebuttal/mnist_lenet5_cva_zoomed_combined.pdf}
%         \caption{MNIST Lenet5}
%     \end{subfigure}%
%     ~
%     \begin{subfigure}[t]{0.49\textwidth}
%         \centering
%         \includegraphics[width=0.48\textwidth]{neurips_rebuttal/mnist_mlp_cva_combined.pdf}
%         \includegraphics[width=0.48\textwidth]{neurips_rebuttal/mnist_mlp_cva_zoomed_combined.pdf}
%         \caption{MNIST MLP}
%     \end{subfigure}%
%     \caption{Accuracy vs. compression for pruning when using different methods on MNIST.}
%     \label{fig:post_training_graphs_mnist}
% \end{figure*}

%\clearpage
%\input{checklist}

\end{document}